\documentclass[runningheads]{llncs}
\usepackage{amsmath, amssymb} 
\usepackage{color}
\usepackage[pagebackref=true,breaklinks=true,letterpaper=true,colorlinks,citecolor=blue,linkcolor=red,bookmarks=false]{hyperref}
\usepackage{booktabs}
\usepackage{graphicx}
\usepackage{mathtools}
\usepackage{multirow}
\usepackage{tabularx}

\begin{document}

\DeclarePairedDelimiter\ceil{\lceil}{\rceil}
\DeclarePairedDelimiter\floor{\lfloor}{\rfloor}
\newcolumntype{Y}{>{\centering\arraybackslash}X}
\pagestyle{headings}
\mainmatter
\def\ECCVSubNumber{798}
\title{GRNet: Gridding Residual Network for \\Dense Point Cloud Completion}

\author{
Haozhe Xie \inst{1,2}\orcidID{0000-0001-9596-5179} \and
Hongxun Yao \inst{1}\orcidID{0000-0003-3298-2574} \and
Shangchen Zhou \inst{3}\orcidID{0000-0001-8201-8877} \and
Jiageng Mao \inst{4}\orcidID{0000-0003-2571-8767} \and
Shengping Zhang \inst{1,5}\orcidID{0000-0001-5200-3420} \and
Wenxiu Sun \inst{2}\orcidID{0000-0001-5026-8820}}
\institute{%
$^1$Harbin Institute of Technology\hspace{0.1in}
$^2$SenseTime Research \\
$^3$Nanyang Technological University\\
$^4$The Chinese University of Hong Kong\hspace{0.1in}
$^5$Peng Cheng Laboratory \\
\url{https://haozhexie.com/project/grnet}}

\titlerunning{GRNet: Gridding Residual Network}
\authorrunning{Haozhe Xie {\it et al.}}



\maketitle

\begin{abstract}
Estimating the complete 3D point cloud from an incomplete one is a key problem in many vision and robotics applications.
Mainstream methods ({\it e.g.,} PCN and TopNet) use Multi-layer Perceptrons (MLPs) to directly process point clouds, which may cause the loss of details because the structural and context of point clouds are not fully considered.
To solve this problem, we introduce 3D grids as intermediate representations to regularize unordered point clouds and propose a novel Gridding Residual Network (GRNet) for point cloud completion.
In particular, we devise two novel differentiable layers, named {\it Gridding} and {\it Gridding Reverse}, to convert between point clouds and 3D grids without losing structural information.
We also present the differentiable {\it Cubic Feature Sampling} layer to extract features of neighboring points, which preserves context information.
In addition, we design a new loss function, namely {\it Gridding Loss}, to calculate the L1 distance between the 3D grids of the predicted and ground truth point clouds, which is helpful to recover details.
Experimental results indicate that the proposed {\it GRNet} performs favorably against state-of-the-art methods on the ShapeNet, Completion3D, and KITTI benchmarks.

\end{abstract}

\section{Introduction}

With the rapid development of 3D acquisition technologies, 3D sensors ({\it e.g.}, LiDARs) are becoming increasingly available and affordable.
As a commonly used format, point clouds are the preferred representation for describing the 3D shape of an object.
Complete 3D shapes are required in many applications, including semantic segmentation and SLAM \cite{DBLP:journals/trob/CadenaCCLSN0L16}.
However, due to limited sensor resolution and occlusion, highly sparse  and incomplete point clouds can be acquired, which causes loss in geometric and semantic information.
Consequently, recovering the complete point clouds from partial observations, named point cloud completion, is very important for practical applications.

In the recent few years, convolutional neural networks (CNNs) have been applied to 2D images and 3D voxels.
Since the convolution can not be directly applied to point clouds due to their irregularity and unorderedness, most of the existing methods \cite{DBLP:conf/cvpr/DaiQN17,DBLP:conf/iccv/HanLHKY17,DBLP:conf/eccv/SharmaGF16,DBLP:conf/cvpr/StutzG18,DBLP:conf/cvpr/NguyenHTPY16,DBLP:conf/iros/VarleyDRRA17,DBLP:conf/nips/LiuTLH19} voxelize the point cloud into binary voxels, where 3D convolutional neural networks can be applied.
However, the voxelization operation leads to an irreversible loss of geometric information.
Other approaches \cite{DBLP:conf/ThreeDim/YuanKHMH18,DBLP:conf/wacv/MandikalR19,DBLP:conf/cvpr/TchapmiKR0S19} use the Multi-Layer Perceptrons (MLPs) to process point clouds directly.
However, these approaches use max pooling to aggregate information across points in a global or hierarchical manner, which do not fully consider the connectivity across points and the context of neighboring points.
More recently, several attempts \cite{DBLP:journals/tog/WangSLSBS19,DBLP:conf/ijcai/Wang0J19} have been made to incorporate graph convolutional networks (GCN) \cite{DBLP:conf/iclr/KipfW17} to build local graphs in the neighborhood of each point in the point cloud.
However, constructing the graph relies on the K-nearest neighbor (KNN) algorithm, which is sensitive to the point cloud density \cite{DBLP:conf/iccv/ThomasQDMGG19}.

Several attempts in point cloud segmentation have been made to capture spatial relationships in point clouds through more general convolution operations.
SPLATNet \cite{DBLP:conf/cvpr/SuJSMK0K18} and InterpConv \cite{DBLP:conf/iccv/MaoWL19} perform convolution on high-dimensional lattices and 3D cubes interpolated from neighboring points, respectively.
However, both of them are based on a strong assumption that the 3D coordinates of the output points are the same as the input points and thus can not be used for 3D point completion.

\begin{figure*}[!t]
  \resizebox{\linewidth}{!} {
    \includegraphics{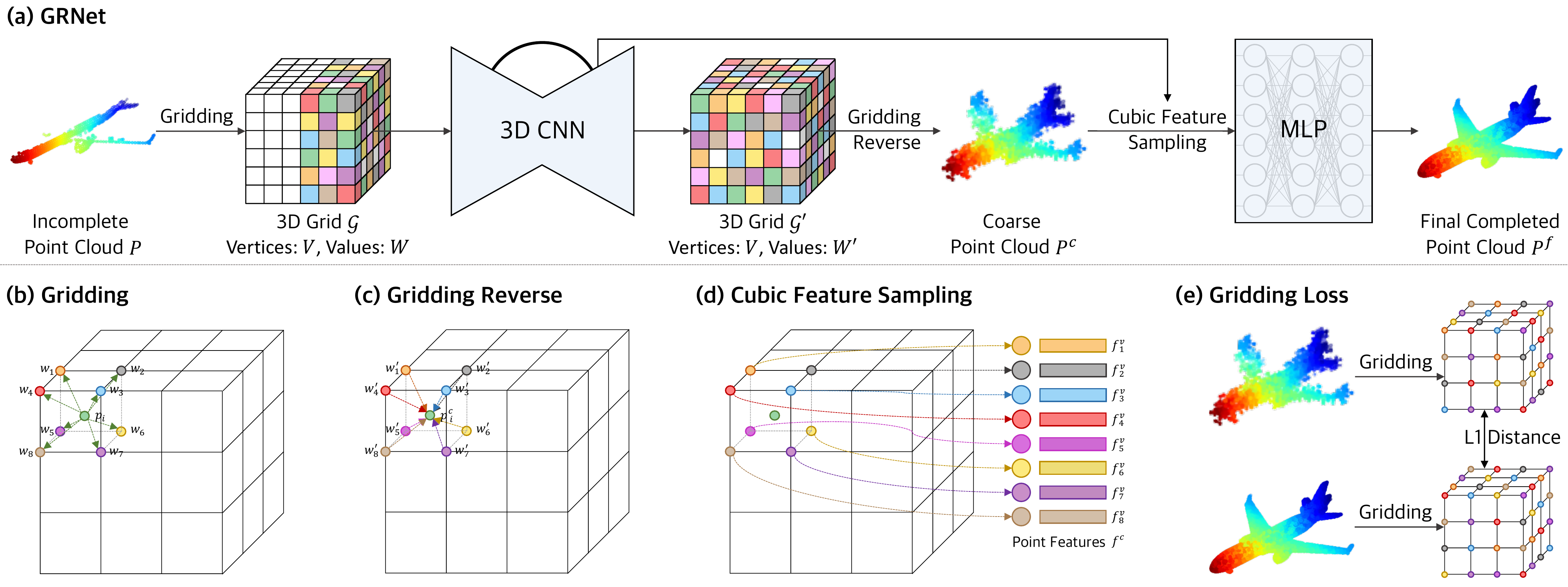}
  }
  \caption{Overview of the proposed {\bf (a)} GRNet, {\bf (b)} Gridding, {\bf (c)} Gridding Reverse, {\bf (d)} Cubic Feature Sampling, and {\bf (e)} Gridding Loss.}
  \label{fig:overview}
\end{figure*}

To address the issues mentioned above, we introduce 3D grids as intermediate representations to regularize unordered point clouds, which explicitly preserves the structural and context of point clouds.
Consequently, we propose a novel Gridding Residual Network ({\it GRNet}) for point cloud completion, as shown in Figure \ref{fig:overview}.
Besides 3D CNN and MLP, we devise three differentiable layers: {\it Gridding}, {\it Gridding Reverse}, and {\it Cubic Feature Sampling}.
In {\it Gridding}, for each point of the point cloud, eight vertices of the 3D grid cell that the point lies in are first weighted using an interpolation function that explicitly measures the geometric relations of the point cloud.
Then, a 3D convolutional neural network (3D CNN) with skip connections is adopted to learn context-aware and spatially-aware features, which allows the network to complete missing parts of the incomplete point cloud.
Next, {\it Gridding Reverse} converts the output 3D grid to a coarse point cloud by replacing each 3D grid cell with a new point whose coordinate is the weighted sum of the eight vertices of the 3D grid cell.
The following {\it Cubic Feature Sampling} extracts features for each point in the coarse point cloud by concatenating the features of the corresponding eight vertices of the 3D grid cell that the point lies in.
The coarse point cloud and the features are forwarded to an MLP to obtain the final completed point cloud. 

Existing methods adopt Chamfer Distance in PSGN \cite{DBLP:conf/cvpr/FanSG17} as the loss function to train the neural networks.
This loss function penalizes the prediction deviating from the ground-truth.
However, there is no guarantee that the predicted point clouds follow the geometric layout of objects, and the networks tend to output a mean shape that minimizes the distance \cite{DBLP:conf/eccv/JiangSQJ18,DBLP:conf/nips/XuWCMN19}.
Some recent works \cite{DBLP:conf/eccv/JiangSQJ18,DBLP:conf/nips/XuWCMN19,DBLP:conf/nips/KarHM17,DBLP:conf/eccv/LiPZR18,DBLP:conf/aaai/LinKL18} attempt to solve the unorderness while preserving fine-grained details by projecting the 3D point cloud to an image, which is then supervised by the corresponding ground truth masks.
However, the projection requires extrinsic camera parameters, which are challenging to estimate in most scenarios \cite{DBLP:conf/cvpr/PengLHZB19}.
To solve the unorderedness of point clouds, we propose {\it Gridding Loss}, which calculates the L1 distance between the generated points and ground truth by representing them in regular 3D grids with the proposed {\it Gridding} layer.

The contributions can be summarized as follows:

\begin{itemize}
  \item We innovatively introduce 3D grids as intermediate representations to regularize unordered point clouds, which explicitly preserve the structural and context of point clouds.
  \item We propose a novel Gridding Residual Network ({\it GRNet}) for point cloud completion. We design three differentiable layers: {\it Gridding}, {\it Gridding Reverse}, and {\it Cubic Feature Sampling}, as well as a new {\it Gridding Loss}.
  \item Extensive experiments are conducted on the ShapeNet, Completion3D, and KITTI benchmarks, which indicate that the proposed {\it GRNet} performs favorably against state-of-the-art methods.
\end{itemize}

\section{Related Work}

According to the network architecture used in point cloud completion and reconstruction, existing networks can be roughly categorized into MLP-based, graph-based, and convolution-based networks.

\noindent \textbf{MLP-based Networks.}
Pioneered by PointNet \cite{DBLP:conf/cvpr/QiSMG17}, several works use MLP for point cloud processing \cite{DBLP:conf/icml/AchlioptasDMG18,DBLP:conf/icmcs/LinXTCD19} and reconstruction \cite{DBLP:conf/ThreeDim/YuanKHMH18,DBLP:conf/wacv/MandikalR19} because of its simplicity and strong representation ability.
These methods model each point independently using several Multi-layer Perceptrons and then aggregate a global feature using a symmetric function ({\it e.g.}, Max Pooling).
However, the geometric relationships among 3D points are not fully considered.
PointNet++ \cite{DBLP:conf/nips/QiYSG17} and TopNet \cite{DBLP:conf/cvpr/TchapmiKR0S19} incorporate a hierarchical architecture to consider the geometric structure.
To relief the structure loss caused by MLP, AtlasNet \cite{DBLP:conf/cvpr/GroueixFKRA18} and MSN \cite{DBLP:conf/aaai/LiuSYSH20} recover the complete point cloud of an object by estimating a collection of parametric surface elements.

\noindent \textbf{Graph-based Networks.}
By considering each point in a point cloud as a vertex of a graph, graph-based networks generate directed edges for the graph based on the neighbors of each point.
In these methods, convolution is usually operated on spatial neighbors, and pooling is used to produce a new coarse graph by aggregating information from each point's neighbors.
Compared with MLP-based methods, graph-based networks take local geometric structures into account.
In DGCNN \cite{DBLP:journals/tog/WangSLSBS19}, a graph is constructed in the feature space and dynamically updated after each layer of the network.
Further, LDGCNN \cite{DBLP:journals/arxiv/1904-10014} removes the transformation network and link the hierarchical features from different layers in DGCNN to improve its performance and reduce the model size.
Inspired by DGCNN, Hassani and Haley \cite{DBLP:conf/iccv/HassaniH19} introduce the multi-scale graph-based network to learn point and shape features for self-supervised classification and reconstruction.
DCG \cite{DBLP:conf/ijcai/Wang0J19} also follows DGCNN to encode additional local connection into a feature vector and progressively evolves from coarse to fine point clouds.

\noindent \textbf{Convolution-based Networks.}
Early works \cite{DBLP:conf/cvpr/DaiQN17,DBLP:conf/iccv/HanLHKY17,DBLP:journals/tvcg/LiSWZ17} usually apply 3D convolutional neural networks (CNNs) build upon the volumetric representation of 3D point clouds.
However, converting point clouds into 3D volumes introduces a quantization effect that discards some details of the data \cite{DBLP:journals/tvcg/WangL19} and is not suitable for representing fine-grained information.
To the best of our knowledge, no work directly applies CNNs on irregular point clouds for shape completion.
In point cloud understanding, several works \cite{DBLP:conf/iccv/MaoWL19,DBLP:conf/cvpr/HuaTY18,DBLP:conf/cvpr/LeiAM19,DBLP:conf/cvpr/LanYYD19,DBLP:conf/nips/LiBSWDC18} develop CNNs operating on discrete 3D grids that are transformed from point clouds.
Hua {\it et al.} \cite{DBLP:conf/cvpr/HuaTY18} define convolutional kernels on regular 3D grids, where the points are assigned with the same weights when falling into the same grid.
PointCNN \cite{DBLP:conf/nips/LiBSWDC18} achieves permutation invariance through a $\chi$-conv transformation.
Besides CNNs on discrete space, several methods \cite{DBLP:conf/nips/LiuTLH19,DBLP:conf/iccv/ThomasQDMGG19,DBLP:conf/cvpr/SuJSMK0K18,DBLP:conf/eccv/XuFXZQ18,DBLP:conf/cvpr/LiuFXP19,DBLP:conf/iccv/LiuFMLXP19,DBLP:conf/cvpr/WuQL19,DBLP:journals/tog/HermosillaRVVR18} define convolutional kernels on continuous space.
Thomas {\it et al.} \cite{DBLP:conf/iccv/ThomasQDMGG19} propose both rigid and deformable kernel point convolution (KPConv) operators for 3D point clouds using a set of learnable kernel points.
Compared with graph-based networks, convolution-based networks are more efficient and robust to point cloud density \cite{DBLP:conf/iccv/MaoWL19}.

\section{Gridding Residual Network}

\subsection{Overview}

The proposed GRNet aims to recover the complete point cloud from an incomplete one in a coarse-to-fine fashion. 
It consists of five components, including {\it Gridding} (Section \ref{sec:gridding}), 3D Convolutional Neural Network  (Section \ref{sec:3dcnn}), {\it Gridding Reverse} (Section \ref{sec:gridding-reverse}), {\it Cubic Feature Sampling} (Section \ref{sec:cubic-feature-sampling}), and Multi-layer Perceptron (Section \ref{sec:mlp}), as shown in Figure \ref{fig:overview}.
Given an incomplete point cloud $P$ as input, {\it Gridding} is first used to obtain a 3D grid $\mathcal{G} = <V,  W>$, where $V$ and $W$ are the vertex set and value set of $\mathcal{G}$, respectively.
Then, $W$ is fed to a 3D CNN, whose output is $W'$. 
Next, {\it Gridding Reverse} produces a coarse point cloud $P^c$ from the 3D grid $\mathcal{G}' = <V,  W'>$.
Subsequently, {\it Cubic Feature Sampling} generates features $F^c$ for the coarse point cloud $P^c$.
Finally, MLP takes the coarse point cloud $P^c$ and the corresponding features $F^c$ as input to produce the final completed point cloud $P^f$.

\subsection{Gridding}
\label{sec:gridding}

2D and 3D convolutions have been developed to process regularly arranged data such as images and voxel grids.
However, it is challenging to directly apply standard 2D and 3D convolutions to unordered and irregular point clouds.
Several methods \cite{DBLP:conf/cvpr/DaiQN17,DBLP:conf/iccv/HanLHKY17,DBLP:conf/nips/LiuTLH19,DBLP:journals/tvcg/LiSWZ17} convert point clouds into 3D voxels and then apply 3D convolutions to them.
However, the voxelization process leads to an irreversible loss of geometric information.
Recent methods \cite{DBLP:conf/ThreeDim/YuanKHMH18,DBLP:conf/cvpr/TchapmiKR0S19} adopt Multi-layer Perceptrons (MLPs) to directly operate on point clouds and aggregate information across points with max pooling.
However, MLP-based methods may lose local context information because the connectivity and layouts of points are not fully considered.
Recent studies also indicate that simply applying MLPs to point clouds cannot always work in practice \cite{DBLP:conf/iccv/MaoWL19,DBLP:conf/eccv/XuFXZQ18}.

In this paper, we introduce 3D grids as intermediate representations to regularize point clouds and further propose a differentiable {\it Gridding} layer, which converts an unordered and irregular point cloud $P = \{p_i\}_{i = 1}^n$ into a regular 3D grid $\mathcal{G} = <V, W>$ while preserving spatial layouts of the point cloud, where $p_i \in \mathbb{R}^3$, $V = \{v_i\}_{i = 1}^{N^3}$, $W = \{w_i\}_{i = 1}^{N^3}$, $v_i \in \{(-\frac{N}{2}, -\frac{N}{2}, -\frac{N}{2}), \dots, (\frac{N}{2} - 1, \frac{N}{2} - 1, \frac{N}{2} - 1)\}$, $w_i \in \mathbb{R}$, $n$ is the number of points in $P$, and $N$ is the resolution of the 3D grid $\mathcal{G}$.
As shown in Figure \ref{fig:overview} (b), we define a cell as a cubic consisting of eight vertices.
For each vertex $v_i = (x_i^v, y_i^v, z_i^v)$ of the 3D grid cell $\mathcal{G}$, we define the neighboring points $\mathcal{N}(v_i)$ as points that lie in the adjacent 8 cells of this vertex.
The point $p = (x, y, z) \in \mathcal{N}(v_i)$ is defined as a neighboring point of vertex $v_i$ by satisfying $p \in P$, $x_i^v - 1 < x < x_i^v + 1$, $y_i^v - 1 < y < y_i^v + 1$, and $z_i^v - 1 < z < z_i^v + 1$, respectively.
In standard voxelization, value $w_i$ at the vertex $v_i$ is computed as
\begin{equation}
  w_i = 
  \begin{cases}
    0 & \forall p \not \in \mathcal{N}(v_i) \\
    1 & \exists p \in \mathcal{N}(v_i) \\
  \end{cases}
\end{equation}
However, this voxelization process introduces a quantization effect that discards some details of an object.
In addition, voxelization is not differentiable and thus can not be applied to point cloud reconstruction.
As illustrated in Figure \ref{fig:overview} (b), given a vertex $v_i$ and its neighboring points $p \in \mathcal{N}(v_i)$, the proposed {\it Gridding} layer computes the corresponding value $w_i$ of this vertex $v_i$ as
\begin{equation}
  w_i = \sum_{p \in \mathcal{N}(v_i)} \frac{w(v_i, p)}{|\mathcal{N}(v_i)|}
\end{equation}
where $|\mathcal{N}(v_i)|$ is the number of neighboring points of $v_i$.
Specially, we define $w_i = 0$ if $|\mathcal{N}(v_i)| = 0$.
The interpolation function $w(v_i, p)$ is defined as
\begin{equation}
  w(v_i, p) = (1 - |x_i^v - x|)(1 - |y_i^v - y|)(1 - |z_i^v - z|)
\end{equation}

\subsection{3D Convolutional Neural Network}
\label{sec:3dcnn}

\begin{figure*}[!t]
  \resizebox{\linewidth}{!} {
    \includegraphics{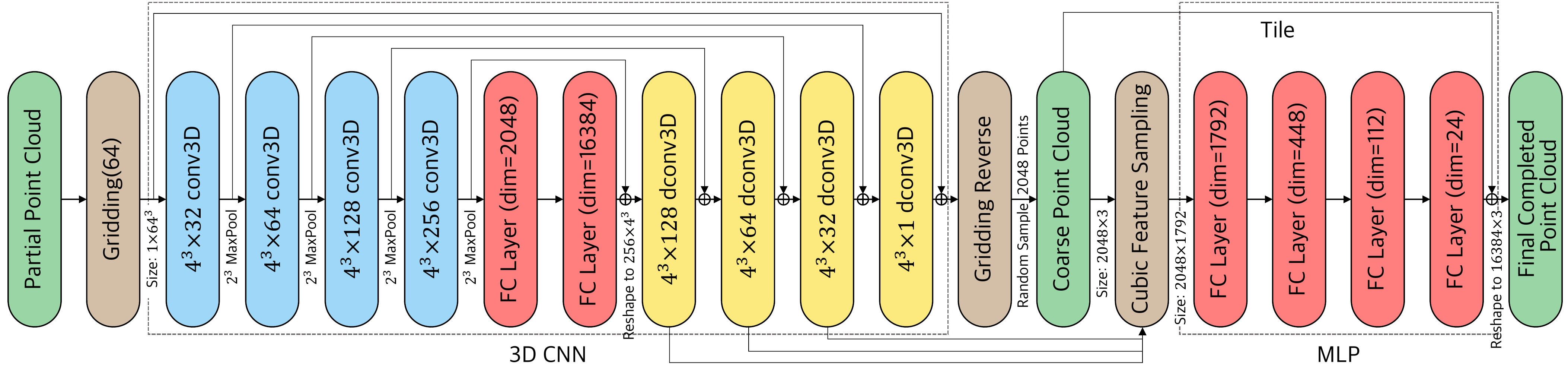}
  }
  \caption{The network architecture of GRNet. $\bigoplus$ denotes the sum operation. Tile creates a new tensor of size $16384 \times 3$ by replicating the ``Coarse Point Cloud'' 8 times.}
  \label{fig:network-architecture}
\end{figure*}

The 3D Convolutional Neural Network (3D CNN) with skip connections aims to complete the missing parts of the incomplete point cloud.
It follows the idea of a 3D encoder-decoder with U-net connections \cite{DBLP:conf/iccv/XieHXSS19,DBLP:journals/ijcv/XieHSSW20}.
Given $W$ as input, the 3D CNN can be formulated as
\begin{equation}
  W' = {\rm 3DCNN}(W)
\end{equation}
where $W' = \{w_i'\}_{i=1}^{N^3}$ and $w_i' \in \mathbb{R}$.

As shown in Figure \ref{fig:network-architecture}, the encoder of the 3D CNN has four 3D convolutional layers, each of which has a bank of $4^3$ filters with padding of 2, followed by batch normalization, leaky ReLU activation, and a max pooling layer with a kernel size of $2^3$.
The numbers of output channels of convolutional layers are $32$, $64$, $128$, $256$, respectively.
The encoder is finally followed by two fully connected layers with dimensions of $2048$ and $16384$.
The decoder consists of four transposed convolutional layers, each of which has a bank of $4^3$ filters with padding of $2$ and stride of $1$, followed by a batch normalization layer and a ReLU activation.

\subsection{Gridding Reverse}
\label{sec:gridding-reverse}

As illustrated in Figure \ref{fig:overview} (c), we propose {\it Gridding Reverse} to generate the coarse point cloud $P^c = \{p_i^c\}_{i=1}^m$ from the 3D grid $\mathcal{G'}=<V, W'>$, where $p_i^c \in \mathbb{R}^3$ and $m$ is the number of points in the coarse point cloud $P^c$.
Let $\Theta^i = \{\theta^i_j\}_{j=1}^8$ be the index set of vertices of the $i-$th 3D grid cell.
{\it Gridding Reverse} generates one point coordinate $p^c_i$ for this grid cell by a weighted combination of eight vertices coordinates $\{v_\theta | \theta \in \Theta^i \}$ and the corresponding values $\{w'_\theta | \theta \in \Theta^i \}$ in this cell, which is computed as
\begin{equation}
  p_i^c = \frac{\sum_{\theta \in \Theta^i} w_\theta' v_\theta}{\sum_{\theta \in \Theta^i} w_\theta'}
\end{equation}
Specially, we ignore the point $p_i^c$ for this cell if $\sum_{\theta \in \Theta^i} w_\theta' = 0$.

\subsection{Cubic Feature Sampling}
\label{sec:cubic-feature-sampling}

MLP-based methods ({\it e.g.}, PCN) are unable to take the context of neighboring points into account due to no local spatial connectivity across points.
These methods use max-pooling to aggregate information globally, which may lose local context information.

To overcome this issue, we present {\it Cubic Feature Sampling} to aggregate features $F^c = \{f^c\}_{i=1}^{m}$ for the coarse point cloud $P^c$, which is helpful for the following MLP to recover the details of point clouds, as shown in Figure \ref{fig:overview} (d).
Let $\mathcal{F} = \{f_1^v, f_2^v, \dots, f_{t^3}^v\}$ be the feature map of 3D CNN, where $f_i^v \in \mathbb{R}^c$ and $t^3$ is the size of the feature map.
For a point $p_i^c$ of the coarse point cloud $P^c$, its features $f_i^c$ are computed as
\begin{equation}
  f_i^c = [f_{\theta_1^i}^v, f_{\theta_2^i}^v, \dots, f_{\theta_8^i}^v]
\end{equation}
where $[\cdot]$ is the concatenation operation.
$\{f_{\theta_j^i}^v\}_{j=1}^8$ denotes the features of eight vertices of the $i$-th 3D gird cell where $p_i^c$ lies in.

In {\it GRNet}, {\it Cubic Feature Sampling} extracts the point features from feature maps generated by the first three transposed convolutional layers in 3D CNN.
To reduce the redundancy of these features and generate a fixed number of points, we randomly sample $2,048$ points from the coarse point cloud $P^c$.
Consequently, it produces a feature map of size $2048 \times 1792$.

\subsection{Multi-layer Perceptron}
\label{sec:mlp}

The Multi-layer Perceptron (MLP) is used to recover the details from the coarse point cloud by learning residual offsets between the coordinates of points in the coarse and final completed point cloud.
It takes the coarse point cloud $P^c$ and the corresponding features $F^c$ as input, and outputs the final completed point cloud $P^f = \{p_i^f\}_{i=1}^k$ as
\begin{equation}
  P^f = {\rm MLP}(F^c) + {\rm Tile}(P^c, r)
\end{equation}
where $p_i^f \in \mathbb{R}^3$ and $k$ is the number of points in the final completed point cloud $P^f$.
Tile creates a new tensor of size $rm \times 3$ by replicating $P^c$ $r$ times.

In {\it GRNet}, $r$ is set to $8$.
The MLP consists of four fully connected layers with dimensions of $1792$, $448$, $112$, and $24$, respectively.
The output of MLP is reshaped to $16384 \times 3$, which corresponds to the offsets of the coordinates of $16,384$ points.

\subsection{Gridding Loss}

Existing methods adopt Chamfer Distance \cite{DBLP:conf/cvpr/FanSG17} as the loss function to train the neural networks.
This loss function penalizes the prediction deviating from the ground-truth.
However, it can not guarantee that the predicted points follow the geometric layout of the object.
Therefore the networks tend to output a mean shape that minimizes the distance, which causes the loss of the object's details \cite{DBLP:conf/eccv/JiangSQJ18,DBLP:conf/nips/XuWCMN19}.

Due to the unorderedness of point clouds, it is difficult to directly apply binary cross-entropy like voxels or L1/L2 loss like images.
With the proposed {\it Gridding}, we can convert unordered point clouds into regular 3D grids (Figure \ref{fig:overview} (e)).
Therefore, we design a new loss function based on {\it Gridding}, namely {\it Gridding Loss}, which is defined as the L1 distance between value sets of the two 3D grids.
Let $\mathcal{G}_{pred} = <V^{pred}, W^{pred}>$ and $\mathcal{G}_{gt} = <V^{gt}, W^{gt}>$ be the 3D grids obtained by {\it Gridding} the predicted and ground truth point clouds, respectively, where $W^{pred} \in \mathbb{R}^{N_G^3}$, $W^{gt} \in \mathbb{R}^{N_G^3}$, and $N_G$ is the resolution of the two 3D grids.
The {\it Gridding Loss} can be defined as
\begin{equation}
  \mathcal{L}_{Gridding}(W^{pred}, W^{gt}) = 
  \frac{1}{N_G^3} \sum ||W^{pred} - W^{gt}||
\end{equation}

\section{Experiments}

\subsection{Datasets}

\noindent \textbf{ShapeNet.}
The ShapeNet dataset \cite{DBLP:conf/cvpr/WuSKYZTX15} for point cloud completion is derived from PCN \cite{DBLP:conf/ThreeDim/YuanKHMH18}, which consists of 30,974 3D models from 8 categories.
The ground truth point clouds containing 16,384 points are uniformly sampled on mesh surfaces.
The partial point clouds are generated by back-projecting 2.5D depth maps into 3D.
For a fair comparison, we use the same train/val/test splits as PCN.

\noindent \textbf{Completion3D.}
The Completion3D benchmark \cite{DBLP:conf/cvpr/TchapmiKR0S19} is composed of 28,974 and 800 samples for training and validation, respectively.
Different from the ShapeNet dataset generated by PCN, there are only 2,048 points in the ground truth point clouds.

\noindent \textbf{KITTI.}
The KITTI dataset \cite{DBLP:journals/ijrr/GeigerLSU13} is composed of a sequence of real-world Velodyne LiDAR scans, also derived from PCN \cite{DBLP:conf/ThreeDim/YuanKHMH18}.
For each frame, the car objects are extracted according to the 3D bounding boxes, which results in 2,401 partial point clouds.
The partial point clouds in KITTI are highly sparse and do not have complete point clouds as ground truth.

\subsection{Evaluation Metrics}

Let $\mathcal{T} = \{(x_i, y_i, z_i)\}_{i=1}^{n_\mathcal{T}}$ be the ground truth and $\mathcal{R} = \{(x_i, y_i, z_i)\}_{i=1}^{n_\mathcal{R}}$ be a reconstructed point set being evaluated, where $n_\mathcal{T}$ and $n_\mathcal{R}$ are the numbers of points of $\mathcal{T}$ and $\mathcal{R}$, respectively.
In our experiments, we use both Chamfer Distance and F-Score as quantitative evaluation metrics.

\noindent \textbf{Chamfer Distance.}
Follow PSGN \cite{DBLP:conf/cvpr/FanSG17} and TopNet \cite{DBLP:conf/cvpr/TchapmiKR0S19}, the distance between $\mathcal{T}$ and $\mathcal{R}$ are defined as
\begin{equation}
  {\rm CD} = \frac{1}{n_\mathcal{T}} \sum_{t \in \mathcal{T}} \min_{r \in \mathcal{R}} ||t - r||^2_2 +
             \frac{1}{n_\mathcal{R}} \sum_{r \in \mathcal{R}} \min_{t \in \mathcal{T}} ||t - r||^2_2
  \label{eq:chamfer-dist}
\end{equation}

\noindent \textbf{F-Score.}
As pointed out in \cite{DBLP:conf/cvpr/TatarchenkoRRLK19}, Chamfer Distance may sometimes be misleading.
As suggested in \cite{DBLP:conf/cvpr/TatarchenkoRRLK19}, we take F-Score as an extra metric to evaluate the  performance of point completion results, which can be defined as following
\begin{equation}
  \textnormal{F-Score}(d) = \frac{2P(d)R(d)}{P(d) + R(d)}
\end{equation}
where $P(d)$ and $R(d)$ denote the precision and recall for a distance threshold $d$, respectively.
\begin{equation}
  P(d) = \frac{1}{n_{\mathcal{R}}} \sum_{r \in \mathcal{R}} \left[\min_{t \in \mathcal{T}} ||t - r|| < d \right]
\end{equation}
\begin{equation}
  R(d) = \frac{1}{n_{\mathcal{T}}} \sum_{t \in \mathcal{T}} \left[\min_{r \in \mathcal{R}} ||t - r|| < d \right]
\end{equation}

\subsection{Implementation Details}

We implement our network using PyTorch \cite{DBLP:conf/nips/AdamSSGEZZALA19} and CUDA\footnote{The source code is available at \url{https://github.com/hzxie/GRNet}.}.
All models are optimized with an Adam optimizer \cite{DBLP:conf/iclr/KingmaB14} with $\beta_1=0.9$ and $\beta_2 = 0.999$.
We train the network with a batch size of $32$ on two NVIDIA TITAN Xp GPUs.
The initial learning rate is set to $1e-4$ and decayed by 2 after 50 epochs.
The optimization is set to stop after 150 epochs.

\subsection{Shape Completion on ShapeNet}

\begin{table*}[!t]
  \caption{Point completion results on ShapeNet compared using Chamfer Distance (CD) with L2 norm computed on 16,384 points and multiplied by $10^4$. The best results are highlighted in bold.}
  \begin{tabularx}{\linewidth}{l|ccYYYYYc|c}
    \toprule
    Methods      & Airplane   & Cabinet    & Car        & Chair 
                 & Lamp       & Sofa       & Table      & Watercraft 
                 & Overall \\
    \midrule  
    AtlasNet \cite{DBLP:conf/cvpr/GroueixFKRA18}    
                 & 1.753      & 5.101      & 3.237      & 5.226
                 & 6.342      & 5.990      & 4.359      & 4.177
                 & 4.523 \\ 
    PCN \cite{DBLP:conf/ThreeDim/YuanKHMH18}
                 & \bf{1.400} & 4.450      & \bf{2.445} & 4.838
                 & 6.238      & 5.129      & 3.569      & 4.062
                 & 4.016 \\
    FoldingNet \cite{DBLP:conf/cvpr/YangFST18}
                 & 3.151      & 7.943      & 4.676      & 9.225
                 & 9.234      & 8.895      & 6.691      & 7.325
                 & 7.142 \\
    TopNet \cite{DBLP:conf/cvpr/TchapmiKR0S19}
                 & 2.152      & 5.623      & 3.513      & 6.346
                 & 7.502      & 6.949      & 4.784      & 4.359
                 & 5.154 \\
    MSN \cite{DBLP:conf/aaai/LiuSYSH20}
                 & 1.543      & 7.249      & 4.711      & 4.539
                 & 6.479      & 5.894      & 3.797      & 3.853
                 & 4.758 \\
    GRNet        & 1.531      & \bf{3.620} & 2.752      & \bf{2.945}
                 & \bf{2.649} & \bf{3.613} & \bf{2.552} & \bf{2.122}
                 & \bf{2.723} \\
  	\bottomrule
  \end{tabularx}
  \label{tab:shapenet-reconstruction-cd}
\end{table*}

\begin{table*}[!t]
  \caption{Point completion results on ShapeNet compared using F-Score@1\%. Note that the F-Score@1\% is computed on 16,384 points. The best results are highlighted in bold.}
  \begin{tabularx}{\linewidth}{l|ccYYYYYc|c}
    \toprule
    Methods      & Airplane   & Cabinet    & Car        & Chair 
                 & Lamp       & Sofa       & Table      & Watercraft 
                 & Overall \\
    \midrule  
    AtlasNet \cite{DBLP:conf/cvpr/GroueixFKRA18}    
                 & 0.845      & 0.552      & 0.630      & 0.552
                 & 0.565      & 0.500      & 0.660      & 0.624
                 & 0.616 \\
    PCN \cite{DBLP:conf/ThreeDim/YuanKHMH18}
                 & 0.881      & \bf{0.651} & \bf{0.725} & 0.625
                 & 0.638      & 0.581      & 0.765      & 0.697
                 & 0.695 \\
    FoldingNet \cite{DBLP:conf/cvpr/YangFST18}
                 & 0.642      & 0.237      & 0.382      & 0.236
                 & 0.219      & 0.197      & 0.361      & 0.299 
                 & 0.322 \\
    TopNet \cite{DBLP:conf/cvpr/TchapmiKR0S19}
                 & 0.771      & 0.404      & 0.544      & 0.413
                 & 0.408      & 0.350      & 0.572      & 0.560
                 & 0.503 \\
    MSN \cite{DBLP:conf/aaai/LiuSYSH20}
                 & \bf{0.885} & 0.644      & 0.665      & 0.657 
                 & 0.699      & 0.604      & \bf{0.782} & 0.708
                 & 0.705 \\
    GRNet        & 0.843      & 0.618      & 0.682      & \bf{0.673}
                 & \bf{0.761} & \bf{0.605} & 0.751      & \bf{0.750} 
                 & \bf{0.708} \\
  	\bottomrule
  \end{tabularx}
  \label{tab:shapenet-reconstruction-fscore}
\end{table*}

\begin{figure*}[!t]
  \resizebox{\linewidth}{!} {
    \includegraphics{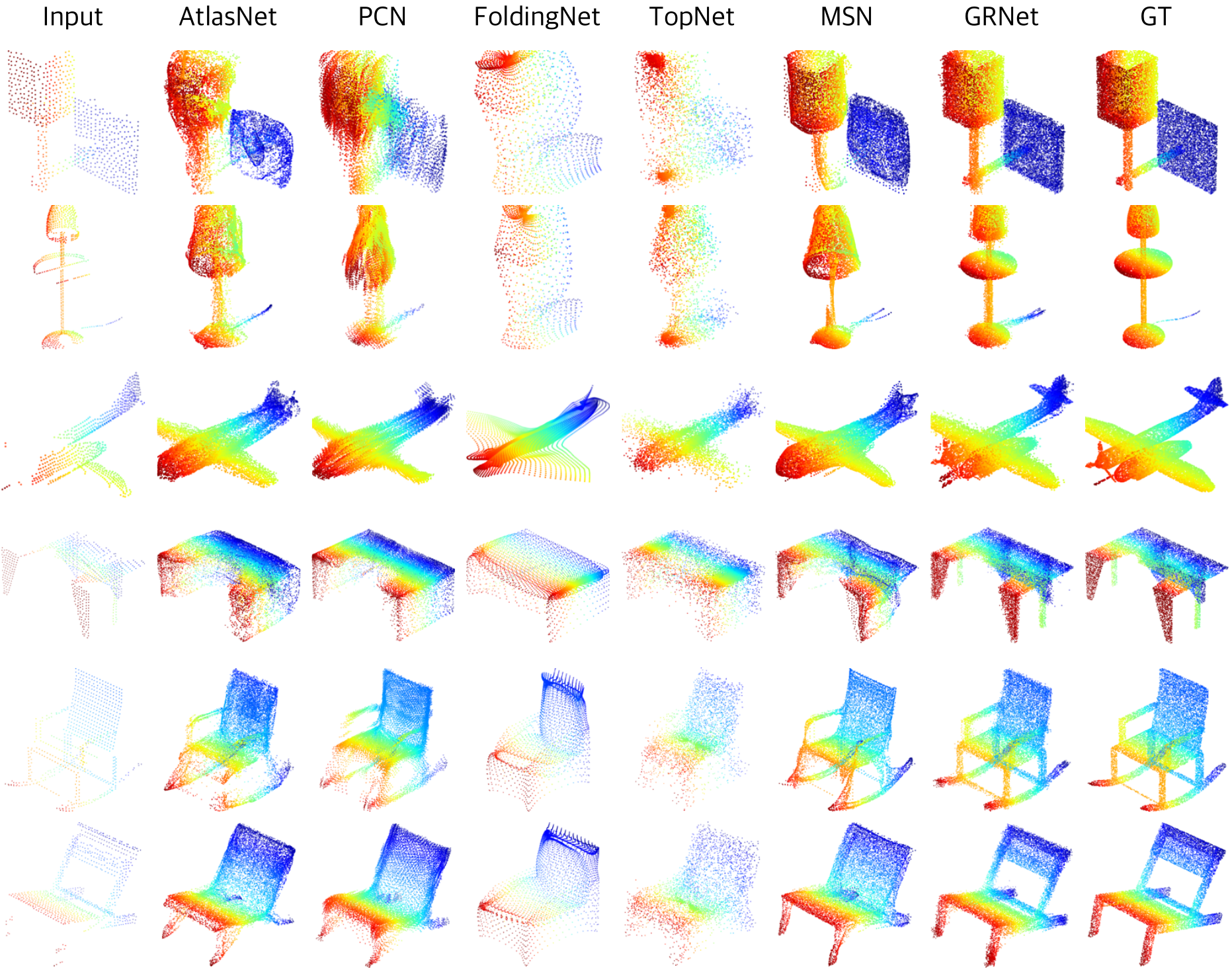}
  }
  \caption{Qualitative completion results on the ShapeNet testing set. GT stands for the ground truth of the 3D object.}
  \label{fig:shapenet-reconstruction}
\end{figure*}

To compare the performance of {\it GRNet} with other state-of-the-art methods, we conduct experiments on the ShapeNet dataset.
\textbf{AtlasNet} \cite{DBLP:conf/cvpr/GroueixFKRA18} generates a point cloud with a set of parametric surface elements.
To compare with other methods fairly, we sample 16,384 points from the generated primitive surface elements.
\textbf{PCN} \cite{DBLP:conf/ThreeDim/YuanKHMH18} completes the partial point cloud with a stacked version of PointNet \cite{DBLP:conf/cvpr/QiSMG17}, which directly outputs the coordinates of 16,384 points.
\textbf{FoldingNet} \cite{DBLP:conf/cvpr/YangFST18} is a baseline method adopted in PCN \cite{DBLP:conf/ThreeDim/YuanKHMH18}, which deforms a $128 \times 128$ 2D grid into 3D point cloud.
\textbf{TopNet} \cite{DBLP:conf/cvpr/TchapmiKR0S19} incorporates a decoder following a hierarchical rooted tree structure to consider the topology of point clouds.
Due to the scalable architecture of TopNet, it can easily generate 16,384 points by setting the number of nodes and the size of feature embedding.
A very recent method \textbf{MSN} \cite{DBLP:conf/aaai/LiuSYSH20} generates dense point cloud containing 8,192 points in a coarse-to-fine fashion.
To generate 16,384 points, we combine the generated points of 2 times forward propagation.

Quantitative results in Tables \ref{tab:shapenet-reconstruction-fscore} and \ref{tab:shapenet-reconstruction-cd} indicate that {\it GRNet} outperforms all competitive methods in terms of Chamfer Distance and F-Score@1\%.
Figure \ref{fig:shapenet-reconstruction} shows the qualitative results for point completion on ShapeNet, which indicates that the proposed method recovers better details of objects ({\it e.g.}, chairs and lamps) than the other methods.

\subsection{Shape Completion on Completion3D}

Using the model with the lowest Chamfer Distance (CD) on the validation set, we recover the complete point clouds for 1,184 objects in the Completion3D testing set.
Then, random subsampling is applied to the generated point clouds to obtain 2,048 points for benchmark evaluation. 
According to the online leaderboard \footnote{\url{https://completion3d.stanford.edu/results}}, as shown in Table \ref{tab:completion-3d-reconstruction}, the overall CD for the proposed {\it GRNet} is $10.64$, which remarkably outperforms state-of-the-art methods and ranks first on this benchmark.

\begin{table*}[!t]
  \caption{Point completion results on Completion3D compared using Chamfer Distance (CD) with L2 norm. Note that the CD is computed on 2,048 points and multiplied by $10^4$. The best results are highlighted in bold.}
  \begin{tabularx}{\linewidth}{l|ccYYYYYc|c}
    \toprule
    Methods      & Airplane   & Cabinet    & Car        & Chair 
                 & Lamp       & Sofa       & Table      & Watercraft 
                 & Overall \\
    \midrule  
    AtlasNet \cite{DBLP:conf/cvpr/GroueixFKRA18}    
                 & 10.36      & 23.40      & 13.40      & 24.16
                 & 20.24      & 20.82      & 17.52      & 11.62
                 & 17.77 \\
    FoldingNet \cite{DBLP:conf/cvpr/YangFST18}
                 & 12.83      & 23.01      & 14.88      & 25.69
                 & 21.79      & 21.31      & 20.71      & 11.51
                 & 19.07 \\
    PCN \cite{DBLP:conf/ThreeDim/YuanKHMH18}
                 & 9.79       & 22.70      & 12.43      & 25.14
                 & 22.72      & 20.26      & 20.27      & 11.73
                 & 18.22 \\
    TopNet \cite{DBLP:conf/cvpr/TchapmiKR0S19}
                 & 7.32       & 18.77      & 12.88      & 19.82
                 & 14.60      & 16.29      & 14.89      & 8.82
                 & 14.25 \\
    GRNet        & \bf{6.13}  & \bf{16.90} & \bf{8.27}  & \bf{12.23}
                 & \bf{10.22} & \bf{14.93} & \bf{10.08} & \bf{5.86}
                 & \bf{10.64} \\
  	\bottomrule
  \end{tabularx}
  \label{tab:completion-3d-reconstruction}
\end{table*}

\subsection{Shape Completion on KITTI}

To evaluate the performance of the proposed method on real-world LiDAR scans, we test {\it GRNet} on the KITTI dataset for completing sparse point clouds of cars.
Unlike ShapeNet generated by back-projected from 2.5D images, point clouds from LiDAR scans can be highly sparse, which are much sparser than those in ShapeNet.

We fine-tuned all competitive methods on ShapeNetCars (the cars from ShapeNet) except PCN that directly uses released output for evaluation.
During testing, each point cloud is transformed into the bounding box's coordinates and transformed back to the world frame after completion.
The models trained specifically on cars are able to incorporate prior knowledge of the object class.

Since there are no complete ground truth point clouds for KITTI, we use Consistency and Uniformity to evaluate the performance of all competitive methods.
Consistency in PCN \cite{DBLP:conf/ThreeDim/YuanKHMH18} is the average CD between the output of the same car instance in $n_f$ consecutive frames.
Let $\mathcal{R}_{t_i}^j$ be the output for the $j$-th car instance at time $t_i$.
The Consistency for the $j$-th car can be calculated as
\begin{equation}
  {\rm Consistency} = \frac{1}{n_f - 1} \sum_{i=2}^{n_f} {\rm CD}(\mathcal{R}_{t_{i - 1}}^j, \mathcal{R}_{t_i}^j)
\end{equation}

\begin{table*}[!t]
  \setlength\tabcolsep{4pt}
  \setlength\extrarowheight{1pt}
  \caption{Point completion results on LiDAR scans from KITTI compared using Consistency and Uniformity. The best results are highlighted in bold.}
  \begin{tabularx}{\linewidth}{l|cYYYYY}
    \toprule
    \multirow{2}{*}{Methods}
             & Consistency
             & \multicolumn{5}{c}{Uniformity for different $p$} \\
             \cline{3-7}
             & ($\times 10^{-3}$)  
             & 0.4\%     & 0.6\%       & 0.8\%      & 1.0\%     
             & 1.2\% \\
    \midrule
    
    AtlasNet \cite{DBLP:conf/cvpr/GroueixFKRA18}    
             & 0.700      & 1.146      & 1.005      & 0.874
             & 0.761      & 0.686\\
    PCN \cite{DBLP:conf/ThreeDim/YuanKHMH18}
             & 1.557      & 3.662      & 5.812      & 7.710
             & 9.331      & 10.823 \\
    FoldingNet \cite{DBLP:conf/cvpr/YangFST18}
             & 1.053      & 1.245      & 1.303      & 1.262
             & 1.162      & 1.063 \\
    TopNet \cite{DBLP:conf/cvpr/TchapmiKR0S19}
             & 0.568      & 1.353      & 1.326      & 1.219
             & 1.073      & 0.950 \\
    MSN \cite{DBLP:conf/aaai/LiuSYSH20}
             & 1.951      & 0.822      & 0.675      & 0.523
             & 0.462      & 0.383 \\
    GRNet    & \bf{0.313} & \bf{0.632} & \bf{0.572} & \bf{0.489}
             & \bf{0.410} & \bf{0.352} \\
    \bottomrule
  \end{tabularx}
  \label{tab:kitti-reconstruction}
\end{table*}

\begin{figure*}[!t]
  \resizebox{\linewidth}{!} {
    \includegraphics{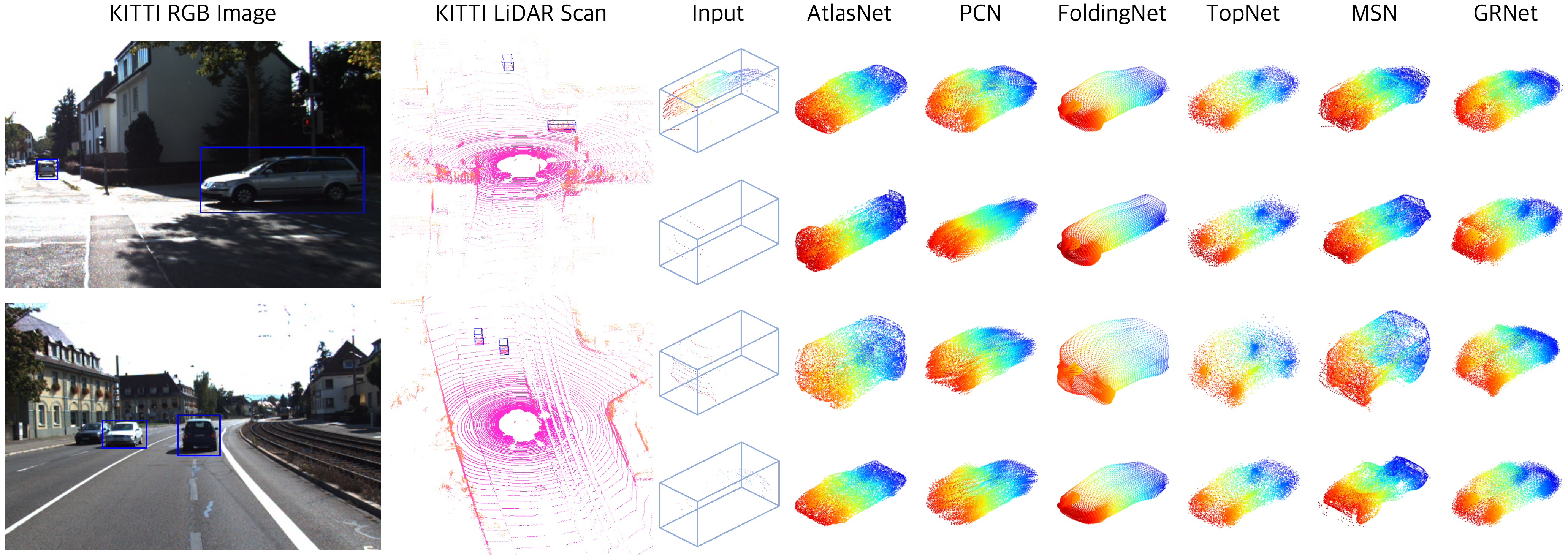}
  }
  \caption{Qualitative completion results on the LiDAR scans from KITTI. The incomplete input point cloud is extracted and normalized from the scene according to its 3D bounding box.}
  \label{fig:kitti-reconstruction}
\end{figure*}

Following PU-GAN \cite{DBLP:conf/iccv/LiLFCH19}, we adopt Uniformity to evaluate the distribution uniformity of the completed point clouds, which can be formulated as
\begin{equation}
  {\rm Uniformity}(p) = \frac{1}{M} \sum_{i=1}^{M} {\rm U_{imbalance}} (S_i){\rm U_{clutter}} (S_i)
\end{equation}
where $S_i (i=1,2,\dots,M)$ is a point subset cropped from a patch of the output $\mathcal{R}$ using the farthest sampling and ball query of radius $\sqrt{p}$.
The term $\rm U_{imbalance}$ and $\rm U_{clutter}$ account for the global and local distribution uniformity, respectively.
\begin{equation}
  {\rm U_{imbalance}}(S_i) = \frac{(|S_i| - \hat{n})^2}{\hat{n}}
\end{equation}
where $\hat{n}=p|\mathcal{R}|$ is the expected number of points in $S_i$.
\begin{equation}
  {\rm U_{clutter}}(S_i) = \frac{1}{|S_i|} \sum_{j = 1}^{|S_i|} \frac{(d_{i, j} - \hat{d})^2}{\hat{d}}
\end{equation}
where $d_{i, j}$ represents the distance to the nearest neighbor for the $j$-th point in $S_i$, and $\hat{d}$ is roughly $\sqrt{\frac{2 \pi p}{|S_i| \sqrt{3}}}$ if $S_i$ has a uniform distribution \cite{DBLP:conf/iccv/LiLFCH19}.

Table \ref{tab:kitti-reconstruction} shows the completion results for cars in the LiDAR scans from the KITTI dataset.
Experimental results indicate that {\it GRNet} outperforms other competitive methods in terms of Consistency and Uniformity.
Benefited from {\it Gridding} and {\it Gridding Reverse}, {\it GRNet} is more sensitive to the spatial structure of the input points, which leads to better consistency between the two consecutive frames.
As shown in Figure \ref{fig:kitti-reconstruction}, the cars are barely recognizable due to incompleteness of the input data.
In contrast, the completed point clouds provide more geometric information.
In addition, the qualitative results also demonstrate the proposed method generates more reasonable shape completion.

\subsection{Ablation Study}

The performance improvement of {\it GRNet} should be attributed to three key components, including {\it Gridding}, {\it Cubic Feature Sampling}, and {\it Gridding Loss}.
To demonstrate the effectiveness of each component in the proposed method, we evaluate the performance with different parameters.

\begin{table*}[!t]
  \setlength\tabcolsep{4pt}
  \setlength\extrarowheight{1pt}
  \caption{The Chamfer Distance (CD), F-Score@1\%, numbers of parameters, and backward time on ShapeNet with different resolutions of 3D grids generated by {\it Gridding}. The backward time is measured on an NVIDIA TITAN Xp GPU with batch size of 1.}
  \begin{tabularx}{1.02\linewidth}{c|cccccc}
    \toprule
    \multirow{2}{*}{Resolutions} 
               & \multicolumn{2}{c}{CD ($\times 10^{-4}$)}
               & \multicolumn{2}{c}{F-Score@1\%}
               & \# Parameters
               & Backward Time \\
               \cline{2-3} \cline{4-5}
               & Coarse      & Complete   & Coarse     & Complete
               & (M)         & (ms) \\
    \midrule
    $32^3$     & 23.339      & 5.943      & 0.329      & 0.549
               & 69.54       & 64 \\
    $64^3$     & \bf{11.259} & \bf{2.723} & 0.340      & 0.708
               & 76.70       & 100 \\
    $128^3$    & 12.383 & 2.732      & \bf{0.366} & \bf{0.712}
               & 76.77       & 302 \\
    \bottomrule
  \end{tabularx}
  \label{tab:ablation-gridding}
\end{table*}

\begin{table*}[!t]
  \setlength\tabcolsep{4pt}
  \setlength\extrarowheight{1pt}
  \caption{The Chamfer Distance (CD), F-Score@1\%, and numbers of parameters of MLPs on ShapeNet with different features maps feeding into {\it Cubic Feature Sampling}. The backward time is measured on an NVIDIA TITAN Xp GPU with batch size of 1.}
  \begin{tabularx}{\linewidth}{YYY|YYcc}
    \toprule
    \multicolumn{3}{c|}{The Size of Feature Maps}
                            & CD          
                            & F-Score
                            & \# Parameters
                            & Backward Time \\
    \cline{1-3}
    $128 \times 8^3$        & 
    $64 \times 16^3$        & 
    $32 \times 32^3$        & $(\times 10^{-4})$      
                            & @1\%       & (M)        & (ms) \\
    \midrule
               &            &            & 11.375     & 0.343
               &  0         & 72 \\
    \midrule
               &            & \checkmark & 2.922      & 0.640
               &  0.11      & 80\\
               & \checkmark & \checkmark & 2.805      & 0.686
               &  0.96      & 88 \\
    \checkmark & \checkmark & \checkmark & \bf{2.723} & \bf{0.708} 
               &  4.07      & 100 \\
    \bottomrule
  \end{tabularx}
  \label{tab:ablation-cubic-feature-sampling}
\end{table*}

\begin{table*}[!t]
  \setlength\tabcolsep{4pt}
  \setlength\extrarowheight{1pt}
  \caption{The Chamfer Distance (CD) and F-Score@1\% on ShapeNet with different resolutions of 3D grids generated by {\it Gridding Loss}. The backward time is measured on an NVIDIA TITAN Xp GPU with batch size of 1.}
  \begin{tabularx}{\linewidth}{c|YYYYc}
    \toprule
    \multirow{2}{*}{Resolutions} 
               & \multicolumn{2}{c}{CD ($\times 10^{-4}$)}
               & \multicolumn{2}{c}{F-Score@1\%}
               & Backward Time \\
               \cline{2-3} \cline{4-5}
               & Coarse      & Complete   & Coarse     & Complete
               & (ms) \\
    \midrule
    Not Used   & 11.259      & 4.460      & 0.340      & 0.624
               & 86 \\
    \midrule
    $64^3$     & 10.275       & 3.427     & 0.364      & 0.672
               & 92 \\
    $128^3$    & \bf{9.324}  & \bf{2.723} & \bf{0.386} & \bf{0.708}
               & 100 \\
    \bottomrule
  \end{tabularx}
  \label{tab:ablation-gridding-loss}
\end{table*}

\noindent \textbf{Gridding.}
Table \ref{tab:ablation-gridding} shows the results of different resolutions of 3D grids generated by {\it Gridding}.
The F-Score of final completed point clouds increases with the 3D grids' resolutions.
However, the numbers of parameters and the backward time also increases.
To archive a balance between effect and efficiency, we choose the resolution of size $64^3$ for {\it Gridding} in {\it GRNet}.

\noindent \textbf{Cubic Feature Sampling.}
To quantitatively evaluate the effect of {\it Cubic Feature Sampling}, we compare the performance without {\it Cubic Feature Sampling} and with different feature maps fed into it.
The experimental results presented in Table \ref{tab:ablation-cubic-feature-sampling} indicate that {\it Cubic Feature Sampling} improves the point cloud completion results significantly.
In addition, with more feature maps are fed, the completion quality becomes better without a significant increase in the numbers of parameters and backward time.

\noindent \textbf{Gridding Loss.}
We further validate the effects of {\it Gridding Loss}, as shown in Table \ref{tab:ablation-gridding-loss}.
There is a decrease in terms of both CD and F-Score when removing  {\it Gridding Loss}.
When increasing the resolution of 3D grids from $64^3$ to $128^3$, there are $25.9\%$ and $5.4\%$ improvements in CD and F-Score, respectively.

\section{Conclusion}

In this paper, we study how to recover the complete 3D point cloud from an incomplete one.
The main motivation of this work is to enable the convolutions on 3D point clouds while preserving their structural and context information.
To this aim, we introduce 3D grids as intermediate representations to regularize unordered point clouds.
We then propose a novel Gridding Residual Network (GRNet) for point cloud completion, which contains three novel differentiable layers: {\it Gridding}, {\it Gridding Reverse}, and {\it Cubic Feature Sampling}, as well as a new {\it Gridding Loss}. 
Extensive comparisons are conducted on the ShapeNet, Completion3D, and KITTI benchmarks, which indicate that the proposed {\it GRNet} performs favorably against state-of-the-art methods.

\noindent \textbf{Acknowledgements.}
This work is supported by the National Natural Science Foundation of China (Nos. 61772158, 61702136 and 61872112), National Key Research and Development Program of China (Nos. 2018YFC0806802 and 2018YFC0832105), and Self-Planned Task (No. SKLRS202002D) of State Key Laboratory of Robotics and System (HIT).

\clearpage
\bibliographystyle{splncs04}
\bibliography{references}

\begin{thebibliography}{10}

\bibitem{DBLP:journals/trob/CadenaCCLSN0L16}
Cadena, C., Carlone, L., Carrillo, H., Latif, Y., Scaramuzza, D., Neira, J.,
  Reid, I.D., Leonard, J.J.:
\newblock Past, present, and future of simultaneous localization and mapping:
  Toward the robust-perception age.
\newblock {IEEE} Transactions on Robotics \textbf{32}(6) (2016)  1309--1332

\bibitem{DBLP:conf/cvpr/DaiQN17}
Dai, A., Qi, C.R., Nie{\ss}ner, M.:
\newblock Shape completion using 3{D}-encoder-predictor {CNN}s and shape
  synthesis.
\newblock In: {CVPR} 2017. (2017)

\bibitem{DBLP:conf/iccv/HanLHKY17}
Han, X., Li, Z., Huang, H., Kalogerakis, E., Yu, Y.:
\newblock High-resolution shape completion using deep neural networks for
  global structure and local geometry inference.
\newblock In: {ICCV} 2017. (2017)

\bibitem{DBLP:conf/eccv/SharmaGF16}
Sharma, A., Grau, O., Fritz, M.:
\newblock {VConv-DAE}: Deep volumetric shape learning without object labels.
\newblock In: {ECCV} 2016 Workshops. (2016)

\bibitem{DBLP:conf/cvpr/StutzG18}
Stutz, D., Geiger, A.:
\newblock Learning 3{D} shape completion from laser scan data with weak
  supervision.
\newblock In: {CVPR} 2018. (2018)

\bibitem{DBLP:conf/cvpr/NguyenHTPY16}
Nguyen, D.T., Hua, B., Tran, M., Pham, Q., Yeung, S.:
\newblock A field model for repairing 3{D} shapes.
\newblock In: {CVPR} 2016. (2016)

\bibitem{DBLP:conf/iros/VarleyDRRA17}
Varley, J., DeChant, C., Richardson, A., Ruales, J., Allen, P.K.:
\newblock Shape completion enabled robotic grasping.
\newblock In: {IROS} 2017. (2017)

\bibitem{DBLP:conf/nips/LiuTLH19}
Liu, Z., Tang, H., Lin, Y., Han, S.:
\newblock Point-voxel {CNN} for efficient 3{D} deep learning.
\newblock In: {NeurIPS} 2019. (2019)

\bibitem{DBLP:conf/ThreeDim/YuanKHMH18}
Yuan, W., Khot, T., Held, D., Mertz, C., Hebert, M.:
\newblock {PCN:} point completion network.
\newblock In: {3{D}V} 2018. (2018)

\bibitem{DBLP:conf/wacv/MandikalR19}
Mandikal, P., Radhakrishnan, V.B.:
\newblock Dense 3{D} point cloud reconstruction using a deep pyramid nxetwork.
\newblock In: {WACV} 2019. (2019)

\bibitem{DBLP:conf/cvpr/TchapmiKR0S19}
Tchapmi, L.P., Kosaraju, V., Rezatofighi, H., Reid, I.D., Savarese, S.:
\newblock Top{N}et: Structural point cloud decoder.
\newblock In: {CVPR} 2019. (2019)

\bibitem{DBLP:journals/tog/WangSLSBS19}
Wang, Y., Sun, Y., Liu, Z., Sarma, S.E., Bronstein, M.M., Solomon, J.M.:
\newblock Dynamic graph {CNN} for learning on point clouds.
\newblock {ACM} Transactions on Graphics \textbf{38}(5) (2019)  146:1--146:12

\bibitem{DBLP:conf/ijcai/Wang0J19}
Wang, K., Chen, K., Jia, K.:
\newblock Deep cascade generation on point sets.
\newblock In: {IJCAI} 2019. (2019)

\bibitem{DBLP:conf/iclr/KipfW17}
Kipf, T.N., Welling, M.:
\newblock Semi-supervised classification with graph convolutional networks.
\newblock In: {ICLR} 2017. (2017)

\bibitem{DBLP:conf/iccv/ThomasQDMGG19}
Thomas, H., Qi, C.R., Deschaud, J., Marcotegui, B., Goulette, F., Guibas, L.J.:
\newblock Kpconv: Flexible and deformable convolution for point clouds.
\newblock In: {ICCV} 2019. (2019)

\bibitem{DBLP:conf/cvpr/SuJSMK0K18}
Su, H., Jampani, V., Sun, D., Maji, S., Kalogerakis, E., Yang, M., Kautz, J.:
\newblock Splatnet: Sparse lattice networks for point cloud processing.
\newblock In: {CVPR} 2018. (2018)

\bibitem{DBLP:conf/iccv/MaoWL19}
Mao, J., Wang, X., Li, H.:
\newblock Interpolated convolutional networks for 3{D} point cloud
  understanding.
\newblock In: {ICCV} 2019. (2019)

\bibitem{DBLP:conf/cvpr/FanSG17}
Fan, H., Su, H., Guibas, L.J.:
\newblock A point set generation network for 3{D} object reconstruction from a
  single image.
\newblock In: {CVPR} 2017. (2017)

\bibitem{DBLP:conf/eccv/JiangSQJ18}
Jiang, L., Shi, S., Qi, X., Jia, J.:
\newblock {GAL:} geometric adversarial loss for single-view 3{D}-object
  reconstruction.
\newblock In: {ECCV} 2018. (2018)

\bibitem{DBLP:conf/nips/XuWCMN19}
Xu, Q., Wang, W., Ceylan, D., Mech, R., Neumann, U.:
\newblock {DISN:} deep implicit surface network for high-quality single-view
  3{D} reconstruction.
\newblock In: NeurIPS 2019. (2019)

\bibitem{DBLP:conf/nips/KarHM17}
Kar, A., H{\"{a}}ne, C., Malik, J.:
\newblock Learning a multi-view stereo machine.
\newblock In: {NIPS} 2017. (2017)

\bibitem{DBLP:conf/eccv/LiPZR18}
Li, K., Pham, T., Zhan, H., Reid, I.D.:
\newblock Efficient dense point cloud object reconstruction using deformation
  vector fields.
\newblock In: {ECCV} 2018. (2018)

\bibitem{DBLP:conf/aaai/LinKL18}
Lin, C., Kong, C., Lucey, S.:
\newblock Learning efficient point cloud generation for dense 3{D} object
  reconstruction.
\newblock In: {AAAI} 2018. (2018)

\bibitem{DBLP:conf/cvpr/PengLHZB19}
Peng, S., Liu, Y., Huang, Q., Zhou, X., Bao, H.:
\newblock Pvnet: Pixel-wise voting network for 6dof pose estimation.
\newblock In: {CVPR} 2019. (2019)

\bibitem{DBLP:conf/cvpr/QiSMG17}
Qi, C.R., Su, H., Mo, K., Guibas, L.J.:
\newblock Point{N}et: Deep learning on point sets for 3{D} classification and
  segmentation.
\newblock In: {CVPR} 2017. (2017)

\bibitem{DBLP:conf/icml/AchlioptasDMG18}
Achlioptas, P., Diamanti, O., Mitliagkas, I., Guibas, L.J.:
\newblock Learning representations and generative models for 3{D} point clouds.
\newblock In: {ICML} 2018. (2018)

\bibitem{DBLP:conf/icmcs/LinXTCD19}
Lin, H., Xiao, Z., Tan, Y., Chao, H., Ding, S.:
\newblock Justlookup: One millisecond deep feature extraction for point clouds
  by lookup tables.
\newblock In: {ICME} 2019. (2019)

\bibitem{DBLP:conf/nips/QiYSG17}
Qi, C.R., Yi, L., Su, H., Guibas, L.J.:
\newblock Point{N}et++: Deep hierarchical feature learning on point sets in a
  metric space.
\newblock In: {NIPS} 2017. (2017)

\bibitem{DBLP:conf/cvpr/GroueixFKRA18}
Groueix, T., Fisher, M., Kim, V.G., Russell, B.C., Aubry, M.:
\newblock A papier-m{\^{a}}ch{\'{e}} approach to learning 3{D} surface
  generation.
\newblock In: {CVPR} 2018. (2018)

\bibitem{DBLP:conf/aaai/LiuSYSH20}
Liu, M., Sheng, L., Yang, S., Shao, J., Hu, S.M.:
\newblock Morphing and sampling network for dense point cloud completion.
\newblock In: {AAAI} 2020. (2020)

\bibitem{DBLP:journals/arxiv/1904-10014}
Zhang, K., Hao, M., Wang, J., de~Silva, C.W., Fu, C.:
\newblock {L}inked {D}ynamic {G}raph {CNN:} learning on point cloud via linking
  hierarchical features.
\newblock arXiv 1904.10014 (2019)

\bibitem{DBLP:conf/iccv/HassaniH19}
Hassani, K., Haley, M.:
\newblock Unsupervised multi-task feature learning on point clouds.
\newblock In: {ICCV} 2019. (2019)

\bibitem{DBLP:journals/tvcg/LiSWZ17}
Li, D., Shao, T., Wu, H., Zhou, K.:
\newblock Shape completion from a single {RGBD} image.
\newblock {IEEE} {T}ransactions on {V}isualization and {C}omputer {G}raphics
  \textbf{23}(7) (2017)  1809--1822

\bibitem{DBLP:journals/tvcg/WangL19}
Wang, Z., Lu, F.:
\newblock {VoxSegNet}: Volumetric {CNN}s for semantic part segmentation of 3{D}
  shapes.
\newblock {IEEE} {T}ransactions on {V}isualization and {C}omputer {G}raphics
  (2019)  DOI: 10.1109/TVCG.2019.2896310

\bibitem{DBLP:conf/cvpr/HuaTY18}
Hua, B., Tran, M., Yeung, S.:
\newblock Pointwise convolutional neural networks.
\newblock In: {CVPR} 2018. (2018)

\bibitem{DBLP:conf/cvpr/LeiAM19}
Lei, H., Akhtar, N., Mian, A.:
\newblock Octree guided {CNN} with spherical kernels for 3{D} point clouds.
\newblock In: {CVPR} 2019. (2019)

\bibitem{DBLP:conf/cvpr/LanYYD19}
Lan, S., Yu, R., Yu, G., Davis, L.S.:
\newblock Modeling local geometric structure of 3{D} point clouds using
  {Geo-CNN}.
\newblock In: {CVPR} 2019. (2019)

\bibitem{DBLP:conf/nips/LiBSWDC18}
Li, Y., Bu, R., Sun, M., Wu, W., Di, X., Chen, B.:
\newblock {PointCNN}: Convolution on x-transformed points.
\newblock In: {NeurIPS} 2018. (2018)

\bibitem{DBLP:conf/eccv/XuFXZQ18}
Xu, Y., Fan, T., Xu, M., Zeng, L., Qiao, Y.:
\newblock {SpiderCNN}: Deep learning on point sets with parameterized
  convolutional filters.
\newblock In: {ECCV} 2018. (2018)

\bibitem{DBLP:conf/cvpr/LiuFXP19}
Liu, Y., Fan, B., Xiang, S., Pan, C.:
\newblock Relation-shape convolutional neural network for point cloud analysis.
\newblock In: {CVPR} 2019. (2019)

\bibitem{DBLP:conf/iccv/LiuFMLXP19}
Liu, Y., Fan, B., Meng, G., Lu, J., Xiang, S., Pan, C.:
\newblock {DensePoint}: Learning densely contextual representation for
  efficient point cloud processing.
\newblock In: {ICCV} 2019. (2019)

\bibitem{DBLP:conf/cvpr/WuQL19}
Wu, W., Qi, Z., Li, F.:
\newblock {PointConv}: Deep convolutional networks on 3{D} point clouds.
\newblock In: {CVPR} 2019. (2019)

\bibitem{DBLP:journals/tog/HermosillaRVVR18}
Hermosilla, P., Ritschel, T., V{\'{a}}zquez, P., Vinacua, A., Ropinski, T.:
\newblock Monte carlo convolution for learning on non-uniformly sampled point
  clouds.
\newblock {ACM} Transactions on Graphics \textbf{37}(6) (2018)  235:1--235:12

\bibitem{DBLP:conf/iccv/XieHXSS19}
Xie, H., Yao, H., Sun, X., Zhou, S., Zhang, S.:
\newblock {Pix2Vox}: Context-aware 3{D} reconstruction from single and
  multi-view images.
\newblock In: {ICCV} 2019. (2019)

\bibitem{DBLP:journals/ijcv/XieHSSW20}
Xie, H., Yao, H., Zhang, S., Zhou, S., Sun, W.:
\newblock {Pix2Vox}++: Multi-scale context-aware 3{D} object reconstruction
  from single and multiple images.
\newblock IJCV DOI 10.1007/s11263-020-01347-6 (2020)

\bibitem{DBLP:conf/cvpr/WuSKYZTX15}
Wu, Z., Song, S., Khosla, A., Yu, F., Zhang, L., Tang, X., Xiao, J.:
\newblock 3{D} {S}hape{N}ets: {A} deep representation for volumetric shapes.
\newblock In: {CVPR} 2015. (2015)

\bibitem{DBLP:journals/ijrr/GeigerLSU13}
Geiger, A., Lenz, P., Stiller, C., Urtasun, R.:
\newblock Vision meets robotics: The {KITTI} dataset.
\newblock International Journal Robotics Research (IJRR) \textbf{32}(11) (2013)
   1231--1237

\bibitem{DBLP:conf/cvpr/TatarchenkoRRLK19}
Tatarchenko, M., Richter, S.R., Ranftl, R., Li, Z., Koltun, V., Brox, T.:
\newblock What do single-view 3{D} reconstruction networks learn?
\newblock In: {CVPR} 2019. (2019)

\bibitem{DBLP:conf/nips/AdamSSGEZZALA19}
Paszke, A., Gross, S., Massa, F., Lerer, A., Bradbury, J., Chanan, G., Killeen,
  T., Lin, Z., Gimelshein, N., Antiga, L., Desmaison, A., Kopf, A., Yang, E.,
  DeVito, Z., Raison, M., Tejani, A., Chilamkurthy, S., Steiner, B., Fang, L.,
  Bai, J., Chintala, S.a.:
\newblock Py{T}orch: An imperative style, high-performance deep learning
  library.
\newblock In: {NeurIPS} 2019. (2019)

\bibitem{DBLP:conf/iclr/KingmaB14}
Kingma, D.P., Ba, J.:
\newblock Adam: {A} method for stochastic optimization.
\newblock In: {ICLR} 2015. (2015)

\bibitem{DBLP:conf/cvpr/YangFST18}
Yang, Y., Feng, C., Shen, Y., Tian, D.:
\newblock Folding{N}et: Point cloud auto-encoder via deep grid deformation.
\newblock In: {CVPR} 2018. (2018)

\bibitem{DBLP:conf/iccv/LiLFCH19}
Li, R., Li, X., Fu, C., Cohen{-}Or, D., Heng, P.:
\newblock {PU-GAN:} a point cloud upsampling adversarial network.
\newblock In: {ICCV} 2019. (2019)

\bibitem{DBLP:conf/iclr/EsserJDRD20}
Esser, S.K., McKinstry, J.L., Bablani, D., Appuswamy, R., Modha, D.S.:
\newblock Learned step size quantization.
\newblock In: {ICLR} 2020. (2020)

\end{thebibliography}

\clearpage
\appendix

In this supplementary material, we provide additional information to complement the manuscript.
First, we present details of {\it Gridding}, {\it Gridding Reverse}, and {\it Cubic Feature Sampling} (Section \ref{sec:more-explanations}).
Second, we provide additional quantitative results on ShapeNet, Completion3D, and KITTI (Sections \ref{sec:shapenet-quantitative-results}, \ref{sec:completion3d-quantitative-results}, and \ref{sec:kitti-quantitative-results}). 
Third, we present additional ablation studies (Section \ref{sec:ablation-studies}).
At last, we present more qualitative results compared to other methods (Section \ref{sec:qualitative-comparisons}).

\section{More Explanations on Gridding, Gridding Reverse, and Cubic Feature Sampling}
\label{sec:more-explanations}

\subsection{Gridding}

According to the manuscript, given a vertex $v_i$ and its neighboring points $p \in \mathcal{N}(v_i)$. 
The proposed {\it Gridding} layer computes the corresponding value $w_i$ of this vertex $v_i$ as

\begin{equation}
  w_i = \sum_{p \in \mathcal{N}(v_i)} \frac{w(v_i, p)}{|\mathcal{N}(v_i)|}
  \label{eq:gridding}
\end{equation}
where $|\mathcal{N}(v_i)|$ is the number of neighboring points of $v_i$ and $w(v_i, p)$ is defined as

\begin{equation}
  w(v_i, p) = (1 - |x_i^v - x|)(1 - |y_i^v - y|)(1 - |z_i^v - z|)
  \label{eq:trilinear-interpolation}
\end{equation}
Based on Equations \ref{eq:gridding} and \ref{eq:trilinear-interpolation}, the partial derivative with respect to $x$ can be calculated as follows

\begin{equation}
  \frac{\partial w_i}{\partial x} = 
  \begin{cases}
  	-\frac{1}{|\mathcal{N}(v_i)|} \sum_{p \in \mathcal{N}(v_i)}(1 - |y_i^v - y|)(1 - |z_i^v - z|), & x > x_i^v \\
  	\frac{1}{|\mathcal{N}(v_i)|} \sum_{p \in \mathcal{N}(v_i)}(1 - |y_i^v - y|)(1 - |z_i^v - z|), & x \le x_i^v \\
  \end{cases}
\end{equation}
where $x$ and $x_i^v$ are the x-coordinates of the point $p$ and vertex $v_i$, respectively.
Similarly, the partial derivative with respect to $y$ and $z$ can be calculated as follows

\begin{equation}
  \frac{\partial w_i}{\partial y} = 
  \begin{cases}
  	-\frac{1}{|\mathcal{N}(v_i)|} \sum_{p \in \mathcal{N}(v_i)}(1 - |x_i^v - x|)(1 - |z_i^v - z|), & y > y_i^v \\
  	\frac{1}{|\mathcal{N}(v_i)|} \sum_{p \in \mathcal{N}(v_i)}(1 - |x_i^v - x|)(1 - |z_i^v - z|), & y \le y_i^v \\
  \end{cases}
\end{equation}

\begin{equation}
  \frac{\partial w_i}{\partial z} = 
  \begin{cases}
  	-\frac{1}{|\mathcal{N}(v_i)|} \sum_{p \in \mathcal{N}(v_i)}(1 - |x_i^v - x|)(1 - |y_i^v - y|), & z > z_i^v \\
  	\frac{1}{|\mathcal{N}(v_i)|} \sum_{p \in \mathcal{N}(v_i)}(1 - |x_i^v - x|)(1 - |y_i^v - y|), & z \le z_i^v \\
  \end{cases}
\end{equation}
where $y$ and $y_i^v$ are the y-coordinates of the point $p$ and vertex $v_i$, respectively.
$z$ and $z_i^v$ are the z-coordinates of the point $p$ and vertex $v_i$, respectively.

\subsection{Gridding Reverse}

\noindent \textbf{Point Coordinates Normalization.}
{\it Gridding Reverse} generates point $p^c_i = (x_i^c, y_i^c, z_i^c)$ for the $i$-th grid cell by a weighted combination of eight vertices $\{v_\theta | \theta \in \Theta^i \}$ and the corresponding values $\{w'_\theta | \theta \in \Theta^i \}$ in this cell, which is calculated as

\begin{equation}
  p_i^c = \frac{\sum_{\theta \in \Theta^i} w_\theta' v_\theta}{\sum_{\theta \in \Theta^i} w_\theta'}
\end{equation}
where $\sum_{\theta \in \Theta^i} w_\theta' \ne 0$ and $\Theta^i = \{\theta^i_j\}_{j=1}^8$ represents the index set of vertices of this 3D grid cell.
Let $(x_\theta^v, y_\theta^v, z_\theta^v)$ be the coordinate of the vertex $v_\theta$, where $x_\theta^v, y_\theta^v, z_\theta^v \in \{-\frac{N}{2}, -\frac{N}{2} + 1, \dots, -\frac{N}{2} - 1\}$ and $N$ is the resolution of the 3D grid.
The x-, y-, and z- coordinates of $p_i^c$ is calculated as

\begin{equation}
  x_i^c = \frac{\sum_{\theta \in \Theta^i} w_\theta' x_\theta^v}{\sum_{\theta \in \Theta^i} w_\theta'}
\end{equation}

\begin{equation}
  y_i^c = \frac{\sum_{\theta \in \Theta^i} w_\theta' y_\theta^v}{\sum_{\theta \in \Theta^i} w_\theta'}
\end{equation}

\begin{equation}
  z_i^c = \frac{\sum_{\theta \in \Theta^i} w_\theta' z_\theta^v}{\sum_{\theta \in \Theta^i} w_\theta'}
\end{equation}
Since the coordinate $(x_i^{gt}, y_i^{gt}, z_i^{gt})$ of the point in the ground truth point cloud satisfies $-1 < x_i^{gt}, y_i^{gt}, z_i^{gt} < 1$.
The coordinates of the point $p_i^c$ are normalized to $(-1, 1)$ by dividing $-\frac{N}{2}$. 

\noindent \textbf{Backward of Gridding Reverse.}
The partial derivative with respect to $w_\theta'$ can be calculated as

\begin{align}
  \frac{\partial x_i^c}{\partial w_\theta'} 
  & = \frac{x_\theta^v}{\sum_{\theta \in \Theta^i} w_\theta'} -
      \frac{\sum_{\theta \in \Theta^i} w_\theta' x_\theta^v}{\left(\sum_{\theta \in \Theta^i} w_\theta'\right)^2} \nonumber \\
  & = \frac{x_\theta^v}{\sum_{\theta \in \Theta^i} w_\theta'} -
      \frac{1}{\sum_{\theta \in \Theta^i} w_\theta'} \cdot x_i^c \nonumber \\
  & = \frac{x_\theta^v - x_i^c}{\sum_{\theta \in \Theta^i} w_\theta'}
\end{align}
Similarly,
\begin{equation}
  \frac{\partial y_i^c}{\partial w_\theta'} = \frac{y_\theta^v - y_i^c}{\sum_{\theta \in \Theta^i} w_\theta'}
\end{equation}

\begin{equation}
  \frac{\partial z_i^c}{\partial w_\theta'} = \frac{z_\theta^v - z_i^c}{\sum_{\theta \in \Theta^i} w_\theta'}
\end{equation}

\subsection{Cubic Feature Sampling}

\noindent \textbf{Point Coordinates Normalization.}
{\it Cubic Feature Sampling} aggregates features $F^c = \{f_i^c\}_{i=1}^{m}$ of the coarse point cloud $P^c = \{p_i^c\}_{i=1}^m$ from the 3D feature map $\mathcal{F} = \{f_i^v\}_{i=1}^{t^3}$, where $f_i^c, f_i^v \in \mathbb{R}^c$, c is the number of channels of $\mathcal{F}$, $m$ is the number of points in the coarse point cloud, and $t$ is the resolution of $\mathcal{F}$.
According to the manuscript, the features $f_i^c$ for $p_i^c = (x_i^c, y_i^c, z_i^c)$ are calculated as

\begin{equation}
  f_i^c = [f_{\theta_1^i}^v, f_{\theta_2^i}^v, \dots, f_{\theta_8^i}^v]
\end{equation}
where $\{f_{\theta_j^i}^v\}_{j=1}^8$ denotes the features of eight vertices of the $i$-th 3D gird cell where $p_i^c$ lies in.
Specifically, the coordinates of the eight vertices $\{(x_{\theta^i_j}^v, y_{\theta^i_j}^v, z_{\theta^i_j}^v)\}_{j=1}^8$ satisfy 
$x_{\theta^i_j}^v \in \{\floor{\frac{t}{2} x_i^c}, \ceil{\frac{t}{2} x_i^c}\}$, 
$y_{\theta^i_j}^v \in \{\floor{\frac{t}{2} y_i^c}, \ceil{\frac{t}{2} y_i^c}\}$, 
and $z_{\theta^i_j}^v \in \{\floor{\frac{t}{2} z_i^c}, \ceil{\frac{t}{2} z_i^c}\}$, respectively.

\noindent \textbf{Backward of Cubic Feature Sampling.}
During backward propagation, the partial derivative with respect to $f_{\theta_j^i}^v$ can be presented as

\begin{equation}
  \frac{\partial f_{i, j}^c}{\partial f_{\theta_j^i}^v} = 1
\end{equation}
where $j \in \{1, 2, \dots, 8\}$ and $f_{i, j}^c$ denotes the $j$-th element in $f_i^c$.

Since $\floor{\cdot}$ and $\ceil{\cdot}$ is not differentiable, the partial derivatives with respect to $x_i^c$, $y_i^c$, and $z_i^c$ are $0$ \cite{DBLP:conf/iclr/EsserJDRD20}, which can be formulated as follows:

\begin{equation}
  \frac{\partial f_{i, j}^c}{\partial x_i^c} = 0
\end{equation}

\begin{equation}
  \frac{\partial f_{i, j}^c}{\partial y_i^c} = 0
\end{equation}

\begin{equation}
  \frac{\partial f_{i, j}^c}{\partial z_i^c} = 0
\end{equation}

\section{Additional Quantitative Results on ShapeNet}
\label{sec:shapenet-quantitative-results}

According to the manuscript, the Chamfer Distance is with L2 norm.
However, PCN \cite{DBLP:conf/ThreeDim/YuanKHMH18} adopts the Chamfer Distance with L1 norm as an evaluation metric, which can be formulated as follows

\begin{equation}
  {\rm CD} = \frac{1}{2} \left( \frac{1}{n_\mathcal{T}} \sum_{t \in \mathcal{T}} \min_{r \in \mathcal{R}} ||t - r|| +
             \frac{1}{n_\mathcal{R}} \sum_{r \in \mathcal{R}} \min_{t \in \mathcal{T}} ||t - r|| \right)
  \label{eq:chamfer-dist}
\end{equation}
where $\mathcal{T} = \{(x_i, y_i, z_i)\}_{i=1}^{n_\mathcal{T}}$ is the ground truth and $\mathcal{R} = \{(x_i, y_i, z_i)\}_{i=1}^{n_\mathcal{R}}$ is the reconstructed point set being evaluated.
$n_\mathcal{T}$ and $n_\mathcal{R}$ are the numbers of points of $\mathcal{T}$ and $\mathcal{R}$, respectively.

Table \ref{tab:shapenet-reconstruction} shows the results of point cloud completion using the Chamfer Distance calculated with Equation \ref{eq:chamfer-dist}.
The values of PCN are \textbf{exactly} the same as Table 4 in the original paper \footnote{\url{https://arxiv.org/pdf/1808.00671}}.

\begin{table*}[!t]
  \caption{Results of point cloud completion on ShapeNet compared using the Chamfer Distance (CD) with L1 norm computed on 16,384 points and multiplied by $10^3$. The best results are highlighted in bold.}
  \begin{tabularx}{1.03\linewidth}{l|cccccccc|c}
    \toprule
    Methods      & Airplane   & Cabinet     & Car        & Chair 
                 & Lamp       & Sofa        & Table      & Watercraft 
                 & Overall \\
    \midrule  
    AtlasNet \cite{DBLP:conf/cvpr/GroueixFKRA18}    
                 & 6.366      & 11.943      & 10.105     & 12.063
                 & 12.369     & 12.990      & 10.331     & 10.607
                 & 10.847 \\ 
    PCN \cite{DBLP:conf/ThreeDim/YuanKHMH18}
                 & \bf{5.502} & 10.625      & \bf{8.696} & 10.998 
                 & 11.339     & 11.676      & 8.590      & 9.665
                 & 9.636 \\
    FoldingNet \cite{DBLP:conf/cvpr/YangFST18}
                 & 9.491      & 15.796      & 12.611     & 15.545
                 & 16.413     & 15.969      & 13.649     & 14.987
                 & 14.308 \\
    TopNet \cite{DBLP:conf/cvpr/TchapmiKR0S19}
                 & 7.614      & 13.311      & 10.898     & 13.823
                 & 14.439     & 14.779      & 11.224     & 11.124
                 & 12.151 \\
    MSN \cite{DBLP:conf/aaai/LiuSYSH20}
                 & 5.596      & 11.963      & 10.776     & 10.620
                 & 10.712     & 11.895      & 8.704      & 9.485
                 & 9.969 \\
    GRNet        & 6.450      & \bf{10.373} & 9.447      & \bf{9.408}
                 & \bf{7.955} & \bf{10.512} & \bf{8.444} & \bf{8.039}
                 & \bf{8.828} \\
  	\bottomrule
  \end{tabularx}
  \label{tab:shapenet-reconstruction}
\end{table*}

\section{Quantitative Results on Completion3D}
\label{sec:completion3d-quantitative-results}

\begin{figure*}[!t]
  \resizebox{\linewidth}{!} {
    \includegraphics{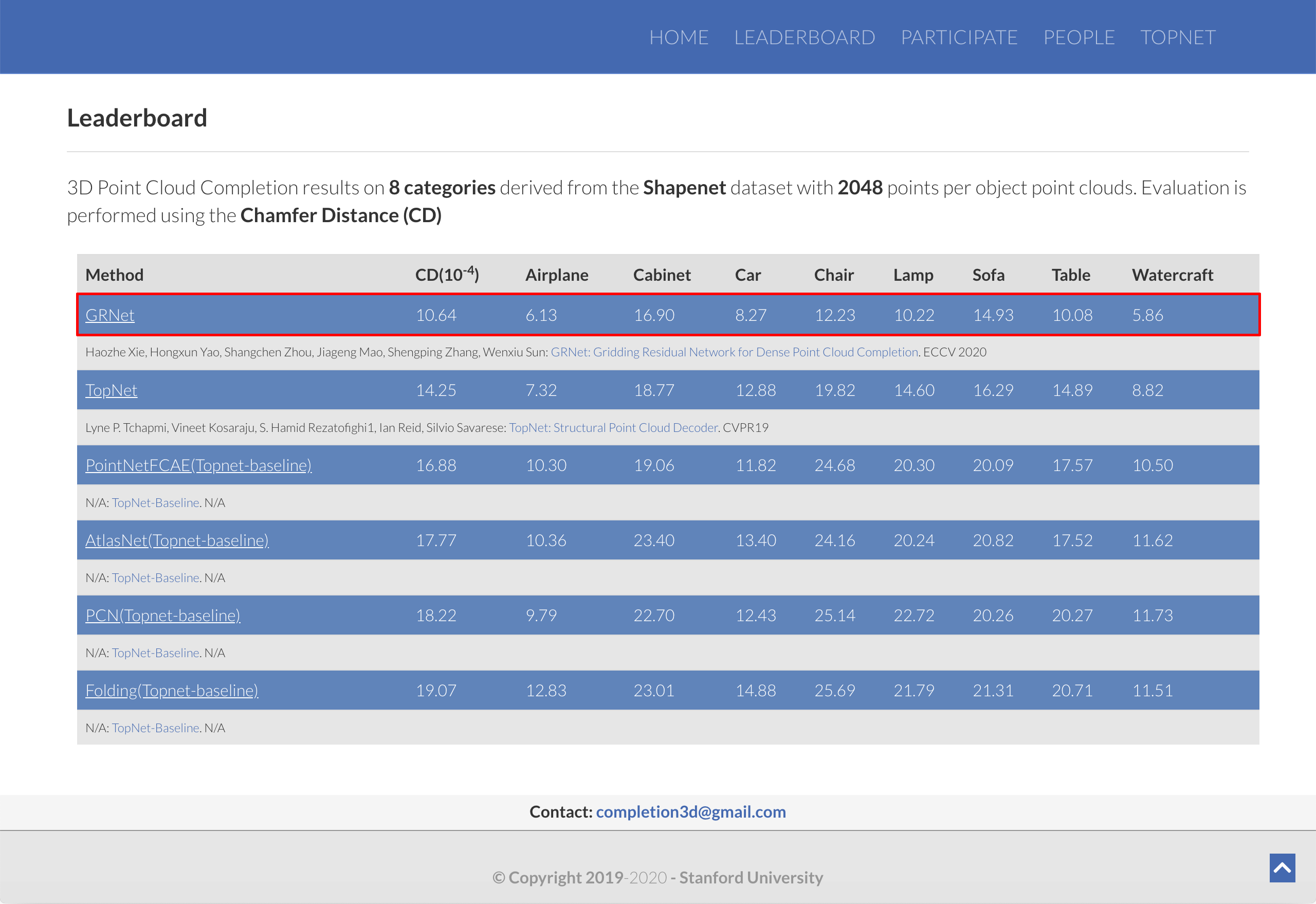}
  }
  \caption{The screenshot of the Completion3D benchmark results. Available online at \url{https://completion3d.stanford.edu/results}}.
  \label{fig:completion-3d-benchmark}
\end{figure*}

Figure \ref{fig:completion-3d-benchmark} is the screenshot of the leaderboard results on the Completion3D benchmark, which is available online at \url{https://completion3d.stanford.edu/results}.

\section{Additional Quantitative Results on KITTI}
\label{sec:kitti-quantitative-results}

\begin{figure*}[!t]
  \resizebox{\linewidth}{!} {
    \includegraphics{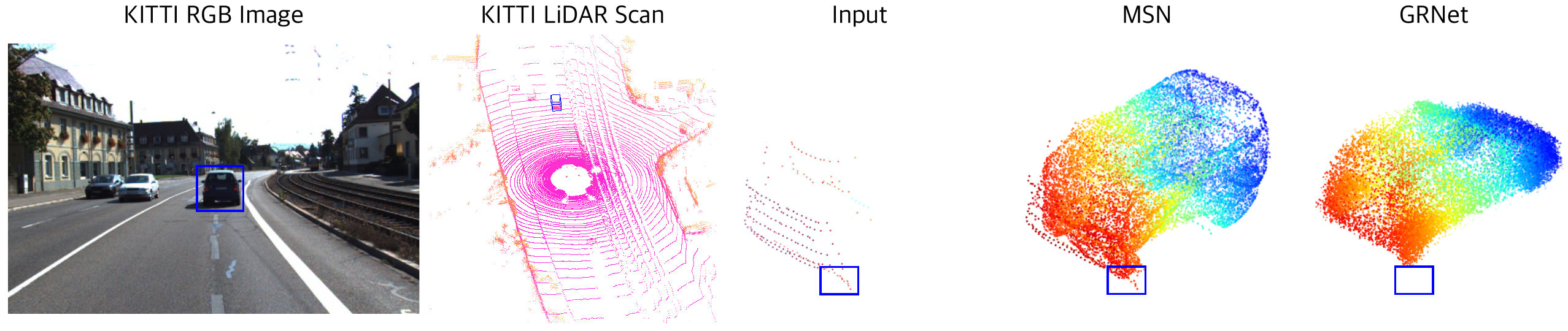}
  }
  \caption{The clutters in the KITTI LiDAR Scan, as shown in the blue bounding box. Compared to MSN, GRNet recovers the complete point cloud while removing the clutters in the input point cloud.}
  \label{fig:kitti-clutters}
\end{figure*}

PCN \cite{DBLP:conf/ThreeDim/YuanKHMH18} uses the Fidelity Distance (FD) and Minimal Matching Distance (MMD) as evaluation metrics for KITTI.
FD is the average distance from each point in the input to its
nearest neighbor in the output, which can be defined as follows

\begin{equation}
 {\rm FD} = \frac{1}{n_\mathcal{I}} \sum_{i \in \mathcal{I}} \min_{r \in \mathcal{R}} ||i - r||_2^2
\end{equation}
where $\mathcal{I}$ denotes the input point cloud.
MMD is the Chamfer Distance (CD) between the output and the car point cloud from ShapeNet that is the closest to the output point cloud in terms of CD.
The Fidelity and MMD on KITTI of the compared methods are shown in Table \ref{tab:kitti-reconstruction}.

However, both FD and MMD are not suitable metrics for KITTI.
As shown in Figure \ref{fig:kitti-clutters}, real-world LiDAR scans usually contain clutters which should be removed in the recovered point cloud.
MSN \cite{DBLP:conf/aaai/LiuSYSH20} incorporates the minimum density sampling (MDS) to preserve the structure of the input point cloud.
Although MSN outperforms other methods in terms of FD, the clutters in the input point cloud are also preserved.
MMD s measures how much the output resembles the cars in ShapeNet.
However, cars from ShapeNet cannot cover all types of cars in the real-world.

\begin{table*}[!t]
  \setlength\tabcolsep{12pt}
  \caption{Results of point cloud completion on KITTI compared using Fidelity Distance (CD) and Minimal Matching Distance (MMD) computed on 16,384 points. Note that both FD and MMD are with L2 norm. The best results are highlighted in bold.}
  \centering
  \begin{tabular}{l|cc}
    \toprule
    Methods      & FD ($\times 10^3$) 
                 & MMD  ($\times 10^3$) \\
    \midrule  
    AtlasNet \cite{DBLP:conf/cvpr/GroueixFKRA18}    
                 & 1.759      & 2.108\\ 
    PCN \cite{DBLP:conf/ThreeDim/YuanKHMH18}
                 & 2.235      & 1.366 \\
    FoldingNet \cite{DBLP:conf/cvpr/YangFST18}
                 & 7.467      & \bf{0.537} \\
    TopNet \cite{DBLP:conf/cvpr/TchapmiKR0S19}
                 & 5.354      & 0.636 \\
    MSN \cite{DBLP:conf/aaai/LiuSYSH20}
                 & \bf{0.434} & 2.259 \\
    GRNet        & 0.816      & 0.568 \\
  	\bottomrule
  \end{tabular}
  \label{tab:kitti-reconstruction}
\end{table*}

\section{Additional Ablation Studies}
\label{sec:ablation-studies}

\noindent \textbf{Number of Sampling Points.}
{\it Gridding Reverse} generates a coarse point cloud from a 3D grid.
We randomly sample 2,048 points from the coarse point cloud to generate a point cloud containing a fixed number of points for the following MLP.
Table \ref{tab:ablation-study-number-of-pts} shows the Chamfer Distance (CD) and F-Score@1\% with different numbers of points sampled.

Experimental results indicate that sampling 2,048 points from the coarse point clouds archives the best performance in terms of CD and F-Score.
The coarse point cloud of an object usually contains about 3,000-4,000 points, oversampling 4,096 points from the coarse point cloud leads to redundant information in the sampled point cloud.
Sampling 1,024 points from the coarse point cloud may lose too much information for the subsequent processing.

\begin{table*}[!t]
  \setlength\tabcolsep{12pt}
  \caption{The Chamfer Distance (CD) and F-Score@1\% on ShapeNet with different numbers of points sampled from the coarse point cloud. The best results are highlighted in bold.}
  \centering
  \begin{tabular}{c|cc}
    \toprule
    \# Points  & CD ($\times 10^{-4}$) & F-Score@1\% \\
    \midrule
    1024       & 2.775                 & 0.697 \\
    2048       & \bf{2.723}            & \bf{0.708} \\
    4096       & 2.832                 & 0.681 \\
    \bottomrule
  \end{tabular}
  \label{tab:ablation-study-number-of-pts}
\end{table*}

\clearpage
\section{Qualitative Comparisons}
\label{sec:qualitative-comparisons}

In this section, we provide more visual comparisons with the state-of-the-art methods \cite{DBLP:conf/ThreeDim/YuanKHMH18,DBLP:conf/cvpr/GroueixFKRA18,DBLP:conf/cvpr/YangFST18,DBLP:conf/cvpr/TchapmiKR0S19,DBLP:conf/aaai/LiuSYSH20} for point cloud completion on ShapeNet \cite{DBLP:conf/cvpr/WuSKYZTX15}.

\vspace{-4 mm}
\begin{figure*}[!h]
  \resizebox{\linewidth}{!} {
    \includegraphics{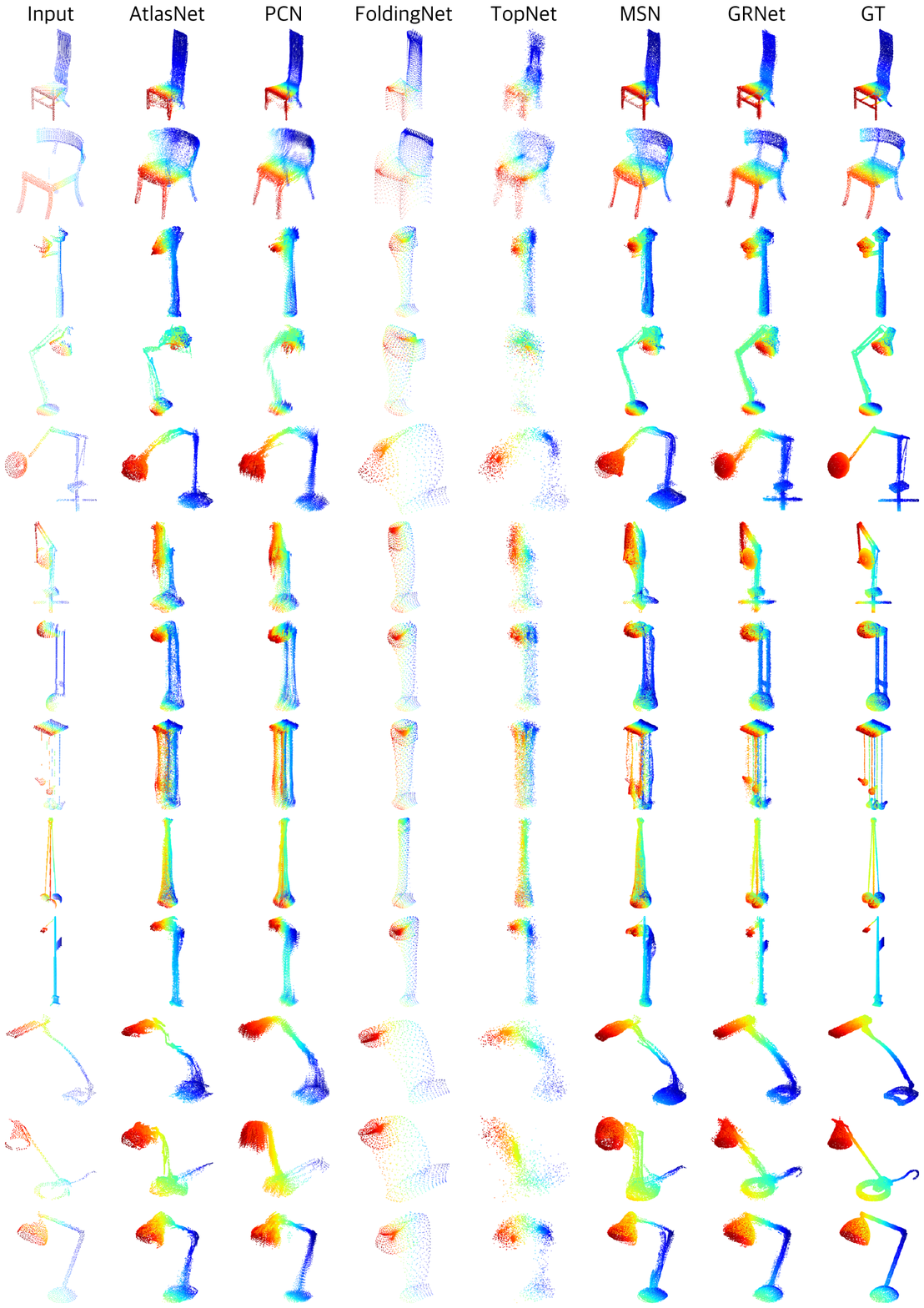}
  }
\end{figure*}

\begin{figure*}[!h]
  \resizebox{\linewidth}{!} {
    \includegraphics{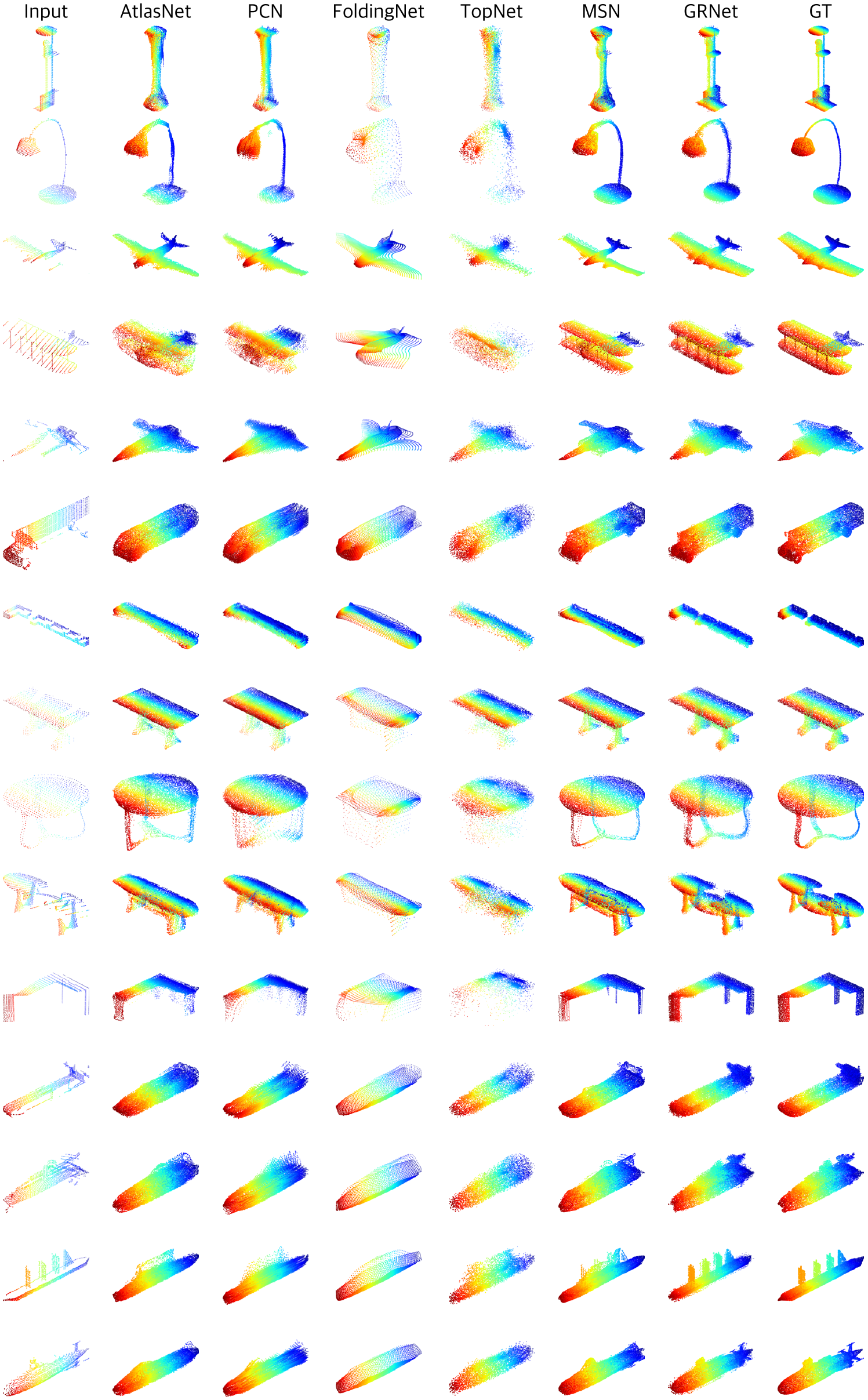}
  }
\end{figure*}

\end{document}